\def\paperID{XXXXX}
\def\confName{CVPR}
\def\confYear{2026}

\def\paperTitle{Improving Diffusion Generalization with Weak-to-Strong Segmented Guidance}

\def\authorBlock{%
\begin{tabular}{c}
Liangyu Yuan\textsuperscript{1,2}\footnotemark[1]\thanks{~Equal contribution. $\dagger$ Corresponding author. This work was done during Liangyu Yuan's visit at WestLake University in 2025. Code for 2d Toy example: \url{https://github.com/851695e35/Leaves_Toy}}
\quad
Yufei Huang$^{1,3}$\footnotemark[1]
\quad
Mingkun Lei$^{1}$
\quad
Tong Zhao$^{1,3}$
\\
Ruoyu Wang$^{1}$
\quad
Changxi Chi$^{1,3}$
\quad
Yiwei Wang$^{4}$
\quad
Chi Zhang$^{1\dagger}$
\\
$^{1}$Westlake University \quad $^{2}$Tongji University \\
$^{3}$Zhejiang University \quad
$^{4}$University of California at Merced \\
\end{tabular}
}

\newif\ifreview 
\newif\ifarxiv \newcommand{\arxiv}{\arxivtrue}
\newif\ifcamera 
\newif\ifrebuttal

\arxiv

\pdfoutput=1
\documentclass[10pt,twocolumn,letterpaper]{article}
\ifreview \usepackage[review]{cvpr} \fi
\ifarxiv \usepackage[pagenumbers]{cvpr} \fi
\ifrebuttal \usepackage[rebuttal]{cvpr} \fi
\ifcamera \usepackage{cvpr} \fi

\usepackage{graphicx}	
\usepackage{amsmath}	
\usepackage{amssymb}	
\usepackage{booktabs}
\usepackage{times}
\usepackage{microtype}
\usepackage{epsfig}
\usepackage{caption}
\usepackage{float}
\usepackage{placeins}
\usepackage{color, colortbl}
\usepackage{stfloats}
\usepackage{enumitem}
\usepackage{tabularx}
\usepackage{xstring}
\usepackage{multirow}
\usepackage{xspace}
\usepackage{url}
\usepackage{subcaption}
\usepackage{xcolor}
\usepackage{algorithmic}
\usepackage{algorithm}
\usepackage[hang,flushmargin]{footmisc}

\ifcamera \usepackage[accsupp]{axessibility} \fi

% ...

\ifarxiv  \fi

\newcommand{\R}[1]{{%
    \textbf{%
        \ifstrequal{#1}{1}{\textcolor{red}{R#1}}{%
        \ifstrequal{#1}{2}{\textcolor{blue}{R#1}}{%
        \ifstrequal{#1}{3}{\textcolor{magenta}{R#1}}{%
        \ifstrequal{#1}{4}{\textcolor{teal}{R#1}}{%
                           \textcolor{cyan}{R#1}%
        }}}}%
    }%
}}

\def\eg{\emph{e.g.}} 

\def\ie{\emph{i.e.}}

\usepackage{xr-hyper}

\makeatletter
\newcommand*{\addFileDependency}[1]{
  \typeout{(#1)}
  \@addtofilelist{#1}
  \IfFileExists{#1}{}{\typeout{No file #1.}}
}

\makeatother
\newcommand*{\myexternaldocument}[1]{
    \externaldocument{#1}
    \addFileDependency{#1.tex}
    \addFileDependency{#1.aux}
}

\definecolor{cvprblue}{rgb}{0.21,0.49,0.74}
\usepackage[pagebackref,breaklinks,colorlinks,allcolors=cvprblue]{hyperref}
\usepackage[capitalize]{cleveref}
\crefname{section}{Sec.}{Secs.}
\crefname{table}{Table}{Tables}
\crefname{figure}{Fig.}{Figs.}

\ifarxiv \crefname{appendix}{App.}{Apps.}
\else \crefname{appendix}{Suppl.}{Suppls.} \fi

\unless\ifarxiv \myexternaldocument{_supplementary} \fi

\begin{document}
\title{\paperTitle}
\author{\authorBlock}
\maketitle

\begin{abstract}
Diffusion models generate synthetic images through an iterative refinement process. However, the misalignment between the simulation-free objective and the iterative process often causes accumulated gradient error along the sampling trajectory, which leads to unsatisfactory results and a failure to generalize. Guidance techniques like Classifier Free Guidance (CFG) and AutoGuidance (AG) alleviate this by extrapolating between the main and inferior signal for stronger generalization. Despite empirical success, the effective operational regimes of prevalent guidance methods are still under-explored, leading to ambiguity when selecting the appropriate guidance method given a precondition. In this work, we first conduct synthetic comparisons to isolate and demonstrate the effective regime of guidance methods represented by CFG and AG from the perspective of weak-to-strong principle. 
Based on this, we propose a hybrid instantiation called SGG under the principle, taking the benefits of both. Furthermore, we demonstrate that the W2S principle along with SGG can be migrated into the training objective, improving the generalization ability of unguided diffusion models. We validate our approach with comprehensive experiments. At inference time, evaluations on SD3 and SD3.5 confirm that SGG outperforms existing training-free guidance variants. Training-time experiments on transformer architectures demonstrate the effective migration and performance gains in both conditional and unconditional settings. Code is available at \url{https://github.com/851695e35/SGG}.
\end{abstract}
\section{Introduction}
\label{sec:intro}

Diffusion and flow matching models have become the de-facto standard for modern image synthesis~\cite{ho2020denoising,sohl2015deep,songscore,lipmanflow,liuflow,albergo2023stochastic,esser2024scaling}, prized for their ability to generate highly realistic images via iterative refinement. However, this multi-step process suffers from a misalignment between the local simulation-free training objective and the global, iterative sampling trajectory. This discrepancy, known as exposure bias~\cite{ning2024elucidating}, leads to the accumulation of network errors during sampling~\cite{Karras2022edm,chen2024GITS}. Consequently, unguided models, particularly for complex conditional generation tasks like text-to-image generation, often fail to generalize properly, producing samples that are out of distribution and perceptually unacceptable~\cite{ho2021classifier}.

To counteract this sampling drift, which is known to degrade generalization~\cite{kadkhodaie2024generalizationdiffusionmodelsarises,song2025selective, li2023generalization}, inference-time guidance has become one of the standard practices, but the effective regimes of different prevalent guidance methods still present ambiguity. Classifier-Free Guidance (CFG)~\cite{ho2021classifier}, for instance, is widely adopted due to its robustness. More recently, AutoGuidance (AG)~\cite{karras2024guiding} was proposed to address a flaw in CFG: the entanglement of condition-adherence and sample diversity. AG attempts to resolve this by guiding the generation with a condition-aligned, inferior model.
However, despite its empirical success on specific scenarios~\cite{karras2024edm2}, the idea of guiding with a condition-aligned weak model has not fully replaced CFG. In complex, large-scale tasks like text-to-image (T2I) generation, AG-inspired methods often serve as a complement to CFG~\cite{rajabi2025token} or are found to be less performant when used in isolation~\cite{liu2025flowing,rajabi2025token}.

\begin{figure*}[tp]
    \centering
    \includegraphics[width=\linewidth]{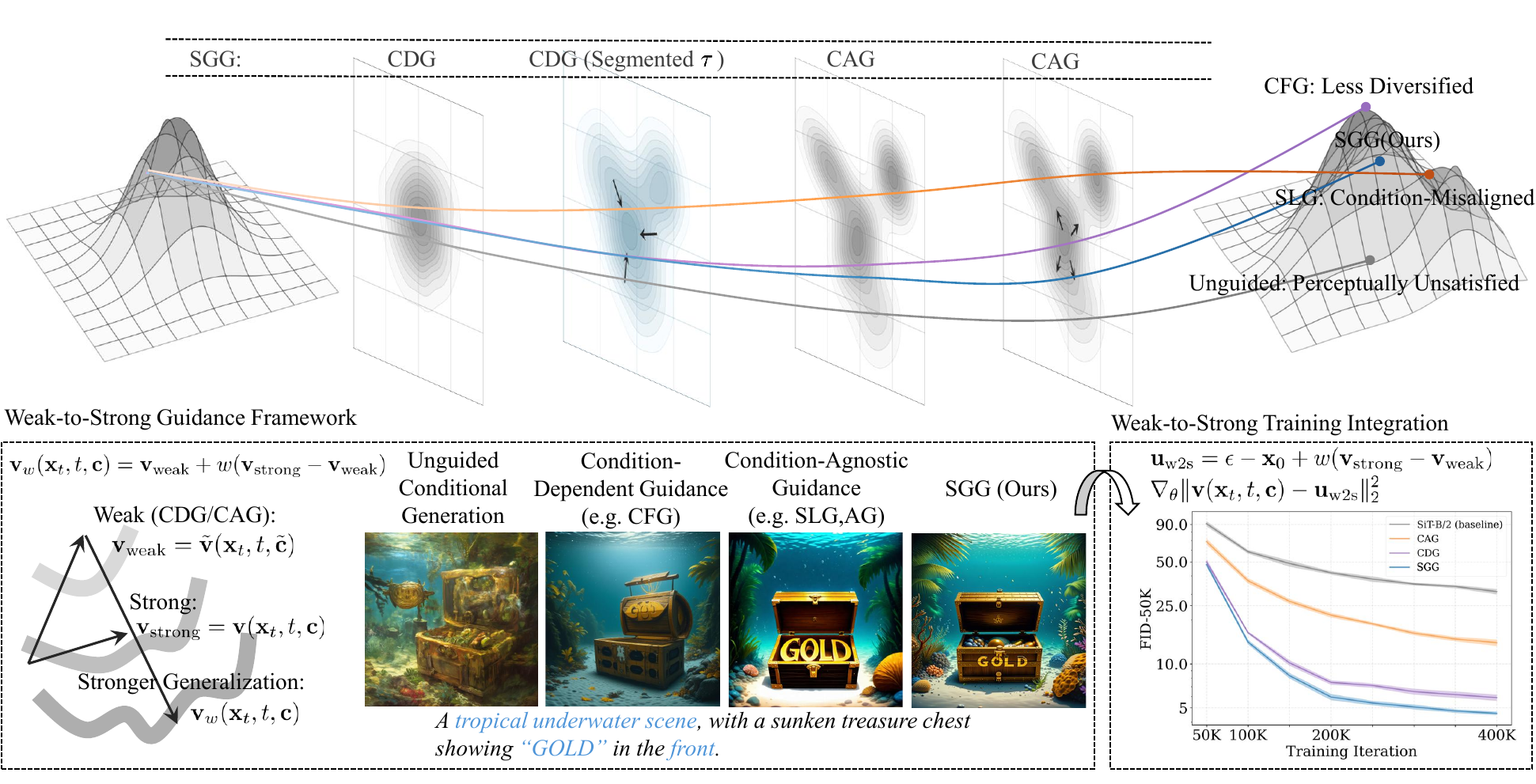}
    \caption{\(\mathrm{I}\): Weak-to-strong guidance principle: Guidance methods serve as tools for improving generalization capacity, we propose SGG to combine the benefits of condition-dependent (CDG) and condition-agnostic guidance (CAG). \(\mathrm{II}\): Integration to the training framework, improving the generalization ability of unguided diffusion models.}
    \label{fig:method}
\end{figure*}

To first give a better understanding of the operational regimes of two types of prevalent guidance methods analogous to CFG~\cite{ho2021classifier} and AG~\cite{karras2024guiding}, we conceptualize them from the perspective of weak-to-strong principle, where we categorize these approaches into two classes: condition-dependent and condition-agnostic. Under this perspective, we conduct synthetic experiments to isolate and demonstrate the effective regimes and failure modes of each class.
Our analysis reveals that appropriate guidance can be influenced by two key factors: the intrinsic \textbf{granularity of the condition}~\cite{zhanconditional} and the \textbf{fitting capacity}~\cite{galashov2025learn,li2023generalization} of the model. Based on this insight, we propose \textbf{S}e\textbf{G}mented \textbf{G}uidance (SGG), a simple yet effective instantiation under the principle that synergizes the benefits of both classes to better handle practical, realistic generation scenarios. Specifically, SGG operates by first leveraging condition-dependent guidance to seek the correct manifold, then switching to condition-agnostic guidance to refine intra-condition details.

We take a step further by migrating the Weak-to-Strong (W2S) guidance principle and SGG from inference directly into the training objective.  This approach enhances the generalization capacity of the unguided diffusion model, thereby reducing the reliance on extra guidance costs during sampling. We also explore various weak-model construction methods, providing a suite of practical choices tailored for transformer architectures. The overall pipeline is illustrated in~\Cref{fig:method}.
We validate our methods in both inference and training settings. For inference, SGG outperforms competing guidance variants on SD3 and SD3.5~\cite{esser2024scaling}. For training, we verify the effectiveness of W2S principle and SGG on SiT models in both conditional and unconditional settings, elevating the generalization capacity of unguided diffusion models. 
Our contribution can be summarized as follows:

\begin{itemize}
    \item We categorizes and analyze the operational regimes of condition-dependent and condition-agnostic guidance under W2S perspective.
    
    \item Based on this analysis, we introduce a hybrid instantiation called SGG, a simple yet effective technique that synergizes the benefits of both guidance paradigms.
    
    \item We migrate W2S principle and SGG from inference-time mechanism into the training objective, directly improving the generation ability of unguided diffusion models.
\end{itemize}

\section{Related work}
\label{sec:related}

\noindent\textbf{Condition-dependent guidance.}
Guidance techniques are crucial for controlling the synthesis process in diffusion models. An early approach, Classifier Guidance (CG)~\cite{dhariwal2021diffusion}, leverages the gradients of a separately trained classifier to steer generation. The now-ubiquitous Classifier-Free Guidance (CFG)~\cite{ho2021classifier} eliminated this need for an external classifier by reformulating the guidance term using Bayes' rule, which requires the model to be jointly trained on conditional and unconditional outputs. Various variants have since been proposed to refine the application of CFG~\cite{Kynkaanniemi2024,fan2025cfgzerostar,sadat2024eliminating,saini2025rectifiedcfgpp,sadatno,gao2025reg,sadatcads,xia2025rectified}. For instance, Guidance Interval~\cite{Kynkaanniemi2024} suggests skipping guidance during specific time intervals to mitigate observed negative effects. APG~\cite{sadat2024eliminating} alleviates the oversaturation problem in high guidance scale through decomposition of the guidance term. 
CFG-Zero-Star~\cite{fan2025cfgzerostar} proposes omitting guidance during the initial sampling steps to enhance performance.

\noindent\textbf{Condition-agnostic guidance}. Recently, the idea of using a condition-aligned inferior model for guidance has emerged in several methods~\cite{karras2024guiding,hyung2025spatiotemporal,ahn2024self,rajabi2025token,wang2025audiomog,chen2025s2guidancestochasticselfguidance,bai2025weak}, serving as either a complement or an alternative to CFG under certain conditions. These methods operate by constructing an inferior prediction to guide the expert output. The inferior signal can be generated in several ways: by training a separate, inferior model, as in AutoGuidance (AG)~\cite{karras2024guiding}. Through self-perturbation, such as skipping residual or attention layers~\cite{hyung2025spatiotemporal,ahn2024self}, by using a stochastic subnetwork, as proposed in $S^2$-Guidance~\cite{chen2025s2guidancestochasticselfguidance}, or by perturbing the input tokens, as in TPG~\cite{rajabi2025token}. However, despite their practical success, these weak-model-based approaches have been reported to be less effective or robust than CFG when used in isolation~\cite{rajabi2025token}, or often function only as a complement to CFG rather than a complete replacement~\cite{chen2025s2guidancestochasticselfguidance}.

\noindent\textbf{Training acceleration in diffusion models}.
Recent works accelerate diffusion model training convergence via two main strategies: improving representation capacities or modifying the regression objective~\cite{yurepresentation,yao2025reconstruction,wang2025diffuse,stoica2025contrastive,xustable,tang2025diffusion,chenvisual}. For representations, REPA~\cite{yurepresentation} aligns the intermediate features of a Diffusion Transformer (DiT) with those from a base model like DINOv2~\cite{oquabdinov2}, VA-VAE~\cite{yao2025reconstruction} applies a similar principle to the features of the tokenizer. SRA~\cite{jiang2025no} propose to align the features of a former block to a latter transformer blocks. For modification on regression objective, contrastive Flow Matching~\cite{stoica2025contrastive} introduces a contrastive term for separation of paths, and STF~\cite{xustable} replace the high-variance single-sample target with a more stable, lower-variance batch-level expectation. GFT~\cite{chenvisual} and MG~\cite{tang2025diffusion} propose to modify the training target by adding the unconditional guidance term from CFG~\cite{ho2021classifier} to enhance generation without guidance.
\section{Preliminaries}
\label{sec:preliminaries}

\noindent\textbf{Diffusion models}. Diffusion Models~\cite{ho2020denoising,sohl2015deep,songscore} are generative models that learn to reverse a process that gradually maps data to noise. Given a predefined data distribution \(p_{\text{data}}\), the general diffusion forward process can be defined by the following perturbation kernel:
\begin{equation}
p(\mathbf{x}_t\mid\mathbf{x}_0)=\mathcal{N}(\mathbf{x}_t; \alpha_t\mathbf{x}_0, \sigma_t^2\mathbf{I}),\quad \mathbf{x}_0\sim p_{\text{data}}
\end{equation}
Where \(\alpha_t,\sigma_t\) defines the noise schedule. 
Under the mathematical equivalence of various noise schedules~\cite{kingma2023understanding,lee2024improving}, we choose the parameterizations of flow matching (\ie stochastic interpolants, \(\alpha_t = 1-t,\sigma_t =t\), along with velocity prediction model)~\cite{lipmanflow,liuflow,albergo2023stochastic} for brevity.
The following reverse time ordinary differential equation is conducted to generate samples:
\begin{equation}
    \mathrm{d}\mathbf{x}_t = \mathbf{v}_{\theta}(\mathbf{x}_t,t)\mathrm{d}t, \mathbf{x}_T\sim\mathcal{N}(0,\mathbf{I}),\quad t: T\to0 
\end{equation}
To obtain the network approximate \(\mathbf{v}_{\theta}(\mathbf{x}_t,t)\) given the state. One has to conduct the following simulation-free conditional flow matching training:
\begin{equation}
\mathbb{E}_{t,\mathbf{x}_t,\mathbf{x}_0,\epsilon}\left[\|v_{\theta}(\mathbf{x}_t,t) - (\epsilon-\mathbf{x}_0)\|^2_2\right]
\end{equation}
Where \(\mathbf{x}_0,\epsilon\) are sampled from the data distribution \(p_{\text{data}}\) and isotropic gaussian \(\mathcal{N}(0,\mathbf{I})\) respectively. The state \(\mathbf{x}_t\) is sampled from the conditional probability path \(p_t(\cdot\mid \mathbf{x}_0,\epsilon)\), which is \(\mathbf{x}_t = (1-t)\mathbf{x}_0+t\epsilon\).

\noindent\textbf{Classifier and Classifier-Free Guidance}. 
To control the generation process, guidance techniques modify the score or velocity field at inference time. Classifier Guidance~\cite{dhariwal2021diffusion} was first introduced to steer generation by sampling from a distribution $p_w(\mathbf{x}_t \mid \mathbf{c}) \propto p(\mathbf{x}_t) p(\mathbf{c} \mid \mathbf{x}_t)^w$, which in practice approximate the score function as:
\begin{equation}
    \nabla_{\mathbf{x}_t}\log p_{w}(\mathbf{x}_t\mid \mathbf{c}) = \nabla_{\mathbf{x}_t} \log p(\mathbf{x}_t) + w \cdot \nabla_{\mathbf{x}_t} \log p(\mathbf{c}\mid\mathbf{x}_t)
\end{equation}
Classifier-Free Guidance (CFG)~\cite{ho2021classifier} avoids the need for a separate classifier by instead training a single diffusion model to learn both conditional and unconditional distributions. This is achieved by randomly dropping the condition $\mathbf{c}$ during training (i.e., replacing it with a null token $\emptyset$). The guided velocity is then formed by extrapolating from the unconditional prediction to the conditional one:
\begin{equation}
    \mathbf{v}_w(\mathbf{x}_t, t, \mathbf{c}) =  w \cdot \mathbf{v}(\mathbf{x}_t, t, \mathbf{c}) + (1-w) \cdot \mathbf{v}(\mathbf{x}_t, t, \emptyset)
\end{equation}
where $w$ is the guidance scale.

\noindent\textbf{Inferior Model Guidance/Condition-Agnostic Guidance}. 
Inferior Model Guidance~\cite{karras2024guiding,hyung2025spatiotemporal,ahn2024self,rajabi2025token,chen2025s2guidancestochasticselfguidance} bears a parameterization-level similarity to CFG but arises from a different motivation. Instead of using an unconditional estimate, it leverages an inferior but condition-aligned model, $\tilde{\mathbf{v}}_{\theta}$, to guide the primary strong model, $\mathbf{v}_{\theta}$. The guided velocity is computed via a similar extrapolation:
\begin{equation}
        \mathbf{v}_w(\mathbf{x}_t, t, \mathbf{c}) =  w \cdot \mathbf{v}(\mathbf{x}_t, t, \mathbf{c}) + (1-w) \cdot \tilde{\mathbf{v}}(\mathbf{x}_t, t, \mathbf{c})
\end{equation}
Here, the weak model $\tilde{\mathbf{v}}$ is conditioned on the same inputs and is designed to be less accurate than $\mathbf{v}_{\theta}$. This is typically achieved by using a smaller network~\cite{karras2024guiding} or by perturbing the architecture of the strong model at inference time~\cite{rajabi2025token,hyung2025spatiotemporal,ahn2024self}. 

\section{Method}
\label{sec:method}

\begin{figure*}[tp]
    \centering
    \includegraphics[width=\linewidth]{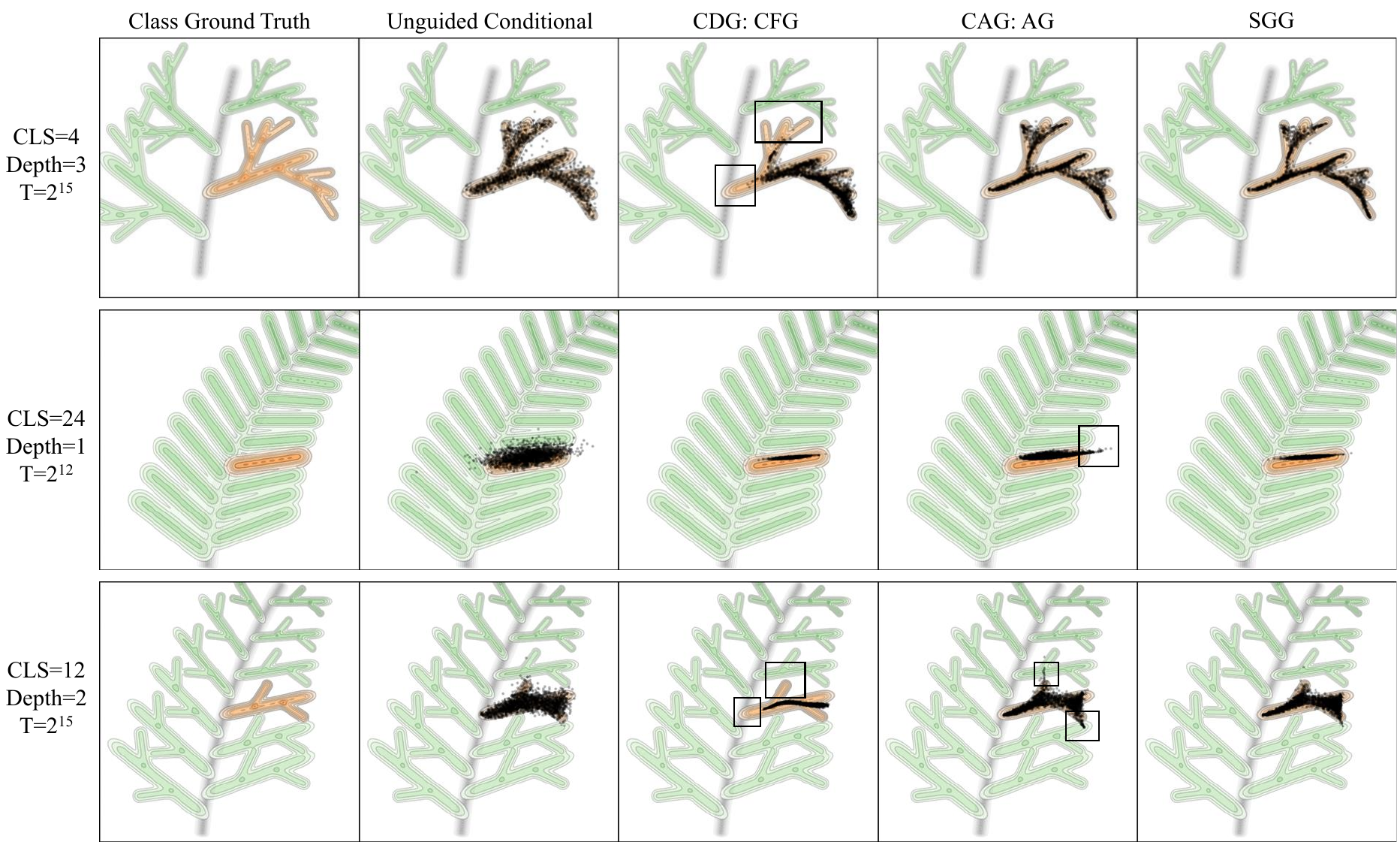}
    \caption{Recursive toy example with varying class complexity and in-class distribution (granular of the condition). 1st row: In a well fitted model and the conditional information is blurry, CFG~\cite{ho2021classifier} exhibits mode-seeking capacity while lack diversity. 2nd row: In a less fitted model and the conditional information is sharp, AG~\cite{karras2024guiding} improves diversity while leads to outliers. 3rd row: In practice, SGG incorporates the mode-seeking capacity of CFG in high noise levels while applying AG in low noise levels to preserve the in-class distribution.}
    \label{fig:toy example}
\end{figure*}


In this section, from the perspective of weak-to-strong (W2S) principle, we first categorize existing guidance methods into two major groups: condition-dependent and condition-agnostic approaches. We analyze the operation regimes of these two categories under various preconditions and, based on our findings, propose SGG that combines their respective benefits. Finally, we extend W2S with SGG beyond inference by migrating it into the training phase, offering a suite of choices to directly improve the unguided diffusion models' generalization capacity.

\subsection{Weak-to-strong guidance principle}

A general extrapolation formula for weak-to-strong guidance can be expressed as:
\begin{align}
\begin{split}
&\mathbf{v}_w(\mathbf{x}_t,t,\mathbf{c}) = \mathbf{v}_{\text{weak}} + w(\mathbf{v}_{\text{strong}}-\mathbf{v}_{\text{weak}}) \\
&\mathbf{v}_{\text{strong}} =\mathbf{v}(\mathbf{x}_t,t,\mathbf{c}), \quad\mathbf{v}_{\text{weak}}=\tilde{\mathbf{v}}(\mathbf{x}_t,t,\tilde{\mathbf{c}})
\end{split}
\end{align}
where $\mathbf{v},
\mathbf{c}$ and $\tilde{\mathbf{v}},\tilde{\mathbf{c}}$ is the strong and weak velocity output and their corresponding condition input. $w$ is the guidance scale. The primary distinction between guidance methods lies in how the weak signal $\tilde{\mathbf{v}}(\mathbf{x}_t,t,\tilde{\mathbf{c}})$ is constructed.

In \textbf{Condition-Dependent Guidance (CDG)}, exemplified by CFG~\cite{ho2021classifier}, creates a weak signal by manipulating the \textit{condition}: the model architecture is identical, but the condition is dropped ($\tilde{\mathbf{v}} = \mathbf{v}, \tilde{\mathbf{c}} = \emptyset$).  On the other hand, \textbf{Condition-Agnostic Guidance (CAG)}~\cite{karras2024guiding,hyung2025spatiotemporal,ahn2024self} creates a weak signal by manipulating the \textit{model}: the condition is preserved (either with or without), but the model itself is made inferior ($\tilde{\mathbf{v}} = \mathbf{v}_{\text{inferior}}, \tilde{\mathbf{c}} = \mathbf{c}$) by either using a separate smaller network~\cite{karras2024guiding} or by perturbing the main model~\cite{hyung2025spatiotemporal,chen2025s2guidancestochasticselfguidance}.




\subsection{Effective regimes of CAG and CDG}

The choice between CAG and CDG is not absolute. On one hand, CAG, such as AG~\cite{karras2024guiding} with EDM2~\cite{karras2024edm2} models, has been shown to outperform CDG (\eg CFG) on class-conditional benchmarks like ImageNet~\cite{karras2024guiding,karras2024edm2}. On the other hand, CDG remains the dominant and more robust method for large-scale text-to-image~\cite{rajabi2025token,ahn2024self} and audio generation~\cite{wang2025audiomog} tasks, where CAG-based methods like AG~\cite{karras2024guiding} and PAG~\cite{ahn2024self} fall short.

While varying factors such as data distribution, training iterations, weak model construction, and sampling initial noise can all contribute to the performance gap, this work investigates the following two perspectives. We interpret that the effectiveness of each guidance type is not absolute, but can be influenced by two key factors: \textbf{granularity of the condition} and the \textbf{model's fitting capacity.}
To visually substantiate this hypothesis, we conduct synthetic experiments across settings to isolate the effective operational regimes of CAG and CDG. For dataset construction, we follow the principle of~\cite{karras2024guiding} by creating a toy dataset based on a recursive mixture of Gaussians. This setup allows us to precisely control the class number (granularity of the condition) and the recursive depth (in-class complexity). We choose CFG and AG as instances of CDG and CAG respectively. Detail of the configuration can be referred in the appendix.

\noindent\textbf{Failure mode of CDG.}
In our first experiment, we simulate a task with conditional ambiguity (\ie, fewer classes) but high in-class complexity: \(\text{CLS}=4,\text{ Depth}=3\). We train the model for $T=2^{15}$ iterations to ensure a relatively strong fit to the in-class distributions.  Illustrated in 1st row of~\Cref{fig:toy example}, we observe that once the model has captured the overall shape of each cluster, applying CDG causes mode-seeking behavior~\cite{dhariwal2021diffusion}: it pushes samples toward high-density regions, failing to cover the lower-density parts of the class manifold. In contrast, CAG avoids mode collapse, sharpening the class distribution while preserving intra-class coverage.
This finding is analogous to results on well-fitted models on ImageNet-1K~\cite{russakovsky2015ImageNet,zheng2025diffusion}, where CFG is inferior to AG~\cite{karras2024edm2} or even under the performance of unguided generation~\cite{zheng2025diffusion}. However, this trend is inverted in large-scale text-to-image generation~\cite{esser2024scaling}. Given the complexity of the task along with the poor unguided generation results, the robustness of CFG consistently surpasses CAG variants like TPG~\cite{rajabi2025token} and PAG~\cite{ahn2024self}.

\begin{figure}[tp]
    \centering
    \includegraphics[width=0.98\linewidth]{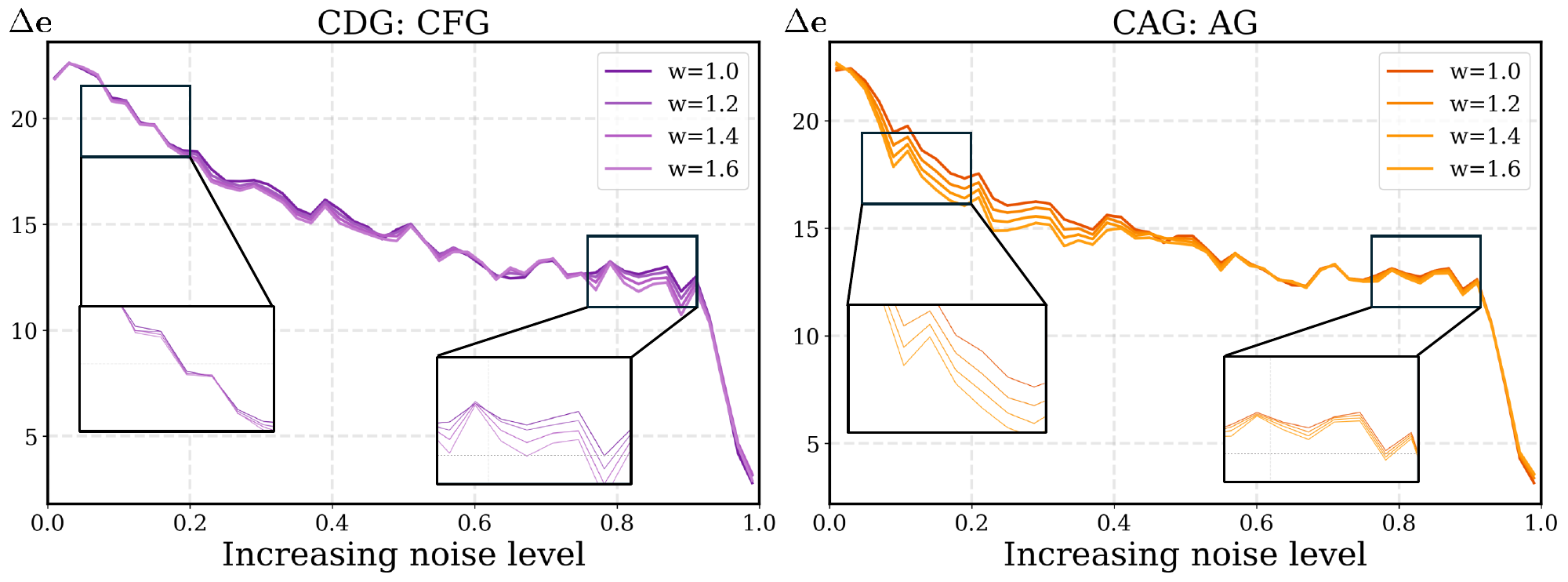}
    \caption{Applying guidance reduces the  gap to optimal velocity $\dot{\mathbf{v}}$. The error-correction of CFG is prominent at high noise levels, while the effect of AG is prominent at low noise levels.}
    \label{fig:lossvsguidance}
\end{figure}

\noindent\textbf{Failure mode of CAG}.
To provide a counter-example where CAG loses its effectiveness in synthetic settings, we now increase the task's conditional complexity while keeping the in-class distribution simple: \(\text{CLS}=24,\text{ Depth}=1\) . We use $T=2^{12}$ training iterations, which is insufficient to fit the data.
As shown in 2nd row of~\Cref{fig:toy example}, the unguided conditional generation from this model produces outliers.  In this underfitted regime, CAG struggles to generate plausible samples and produces artifacts that lie off-manifold or belong to incorrect classes. CDG, in contrast, successfully mitigates this failure by strongly enforcing the condition, it steers the errant samples back toward their classes, removing outliers.
We therefore infer that CDG excels at \textbf{inter-class separation} and class manifold seeking. And CAG is better suited for \textbf{intra-class refinement} once the model is already well-fitted to the condition manifolds.

\noindent\textbf{Simulating realistic scenarios}.
Practical applications, such as large-scale text-to-image models, are characterized by complex condition and detailed in-class distributions. Usually needs large guidance scale (\eg 7.5 for Stable Diffusion~\cite{esser2024scaling, rombach2022high}) for better generalization. We now increase the recursive depth to 2 with 12 classes. In this setting, the model (\(T=2^{15}\)) captures the approximate in-class shape but still produces significant outliers. Illustrated in 3rd row of~\Cref{fig:toy example}, both standard guidance methods fail in distinct ways: CDG, as before, exhibits mode-seeking behavior~\cite{dhariwal2021diffusion,ho2021classifier} and collapses the in-class structure, while CAG preserves the general shape but fails to correct the outliers, leaving them far from the data manifold. Given the trade-offs of CAG and CDG, it is natural to take a step further by devising a practical implementation based on their operational regimes.

\noindent\textbf{Introducing Segmented Guidance (SGG)}.
To bridge the gap between our 2D synthetic analysis and high-dimensional images, we now quantify the error-correction capacities of CDG and CAG on ImageNet~\cite{russakovsky2015ImageNet}. This allows us to investigate their operational regimes in a realistic, high-dimensional setting.

Theoretically, a perfectly fitted model could reconstruct the entire training set (memorization), but in practice, network inductive biases and inevitable approximation error lead to generalization~\cite{gumemorization}. In large-scale tasks~\cite{russakovsky2015ImageNet,esser2024scaling}, this approximation error accumulates, often causing the unguided model's trajectory to drift far from the data manifold and resulting in perceptually unsatisfying samples. To understand how CDG and CAG alleviate this, we pretrain a SiT-B/2 model on ImageNet. We then compute the guided velocity $\mathbf{v}_w(\mathbf{x}_t,t,\mathbf{c})$ for both CFG (as CDG) and AG (as CAG) across all timesteps during generation and measure its distance to the theoretical optimal velocity, $\dot{\mathbf{v}}(\mathbf{x}_t,t, \mathbf{c})$.

The optimal conditional velocity $\dot{\mathbf{v}}$, derived from the dataset (please refer to appendix for derivation and configuration), is:
\begin{align}
\dot{\mathbf{v}}(\mathbf{x}_t,t, \mathbf{c}) &= \mathbb{E}_{\mathbf{x}_0 \sim p(\cdot|\mathbf{c}),\epsilon\sim \mathcal{N}(0,\mathbf{I})}[\mathbf{u}=\boldsymbol{\epsilon}-\mathbf{x}_0 \mid \mathbf{x}_t, t] \\
&=\frac{\sum_{i=1}^N(\mathbf{x}_t-\mathbf{x}_0^i)\mathcal{N}(\mathbf{x}_t;(1-t)\mathbf{x}_0^i,t^2\mathbf{I})}{t\sum_{j=1}^N\mathcal{N}(\mathbf{x}_t;(1-t)\mathbf{x}_0^j,t^2\mathbf{I})}
\end{align}
We measure the guidance error as the Inception distance~\cite{heusel2017gans} between the guided and optimal velocities, $\Delta \mathbf{e}=\mathbb{E}_{\mathbf{x}_t}[\mathrm{d}(\dot{\mathbf{v}} , \mathbf{v}_w)]$, capturing the perceptual alignment on high dimension images. As observed in~\Cref{fig:lossvsguidance}, the error-correction properties of the two classes are temporally separated: CDG (CFG) is most effective at high noise levels, while CAG (AG) is more effective at low noise levels. This corroborates the finding that semantic, high-level information (inter-class) is resolved in early sampling steps~\cite{FuTCFG2025,zhanconditional}, while fine-grained perceptual details (intra-class) are resolved in late steps close to data~\cite{chen2024GITS}.

Inspired by these distinct operational regimes, we propose a simple yet effective hybrid mechanism called \textbf{SGG} (\underline{Se}gmented \underline{G}uidance). Formally, the guided velocity $\mathbf{v}_{w}$ is:
\begin{align}
\label{eq:segmented_guidance_main}
    \mathbf{v}_{w}(\mathbf{x}_t, t, \mathbf{c}) = \mathbf{v}(\mathbf{x}_t, t, \mathbf{c}) + (w-1)\cdot\mathbf{g}(\mathbf{x}_t, t, \mathbf{c})
\end{align}
where $\mathbf{v}$ is the strong model, $w$ is the guidance scale, and the guidance direction $\mathbf{g}$ is segmented by time $\tau$:
\begin{align}
\label{eq:segmented_guidance_term}
    \mathbf{g}(\mathbf{x}_t, t, \mathbf{c}) = \begin{cases}
             \mathbf{v}(\mathbf{x}_t, t, \mathbf{c}) - \mathbf{v}(\mathbf{x}_t, t, \emptyset) & \text{if } t >\tau \\
             \mathbf{v}(\mathbf{x}_t, t, \mathbf{c}) - \tilde{\mathbf{v}}(\mathbf{x}_t, t, \mathbf{c})   & \text{if } t \leq \tau
        \end{cases}
\end{align}
The core idea is to first leverage CDG for \textbf{condition manifold seeking} at high noise levels ($t \ge \tau$) and subsequently apply CAG for \textbf{in-condition refinement} at low noise levels ($t < \tau$).

\subsection{Training integration}

While both the regression target and guidance mechanisms are critical for generalization~\cite{song2025selective,karras2024guiding}, they remain fundamentally decoupled, applied separately during training and sampling. We take a step forward by integrating the Weak-to-Strong (W2S) principle with SGG directly into the training phase. This approach aims to improve the generalization of unguided diffusion model, thereby boosting inference efficiency by reducing  the need for an extra guidance call.

\begin{figure*}[tp]
    \centering
    \includegraphics[width=0.98\linewidth]{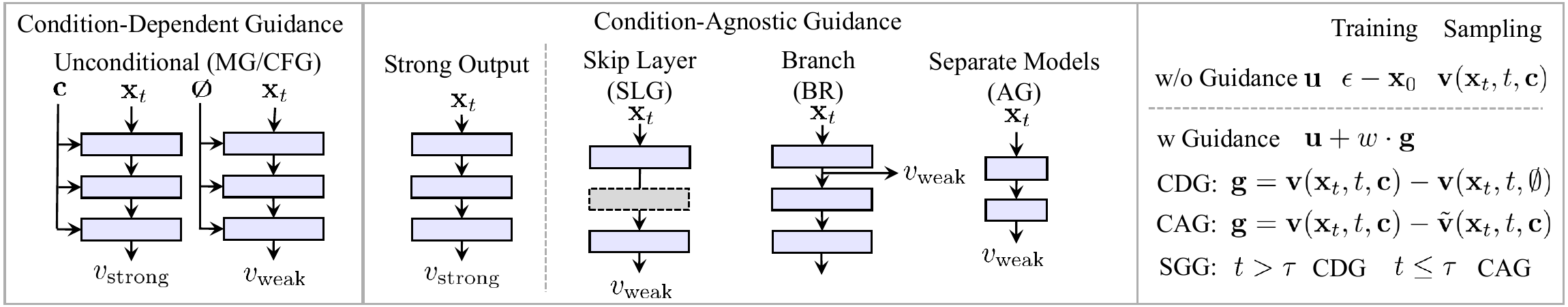}
    \caption{\(\mathrm{I}\): Two groups of construction of the weak models, condition-dependent and condition-agnostic. \(\mathrm{II}\): Segmented Guidance applied in training and sampling}
    \label{fig:arch}
    \vspace{-0.4cm}
\end{figure*}

\noindent\textbf{Training target modification.}
To integrate the extrapolation capacity of guidance explicitely into the training phase, thus reducing the extra forward call of guidance during inference, we modify the standard velocity-matching objective. The conventional training target is the coupling level~\cite{lipmanflow} optimal transport $\mathbf{u} = \boldsymbol{\epsilon} - \mathbf{x}_0$. We augment this target with a guidance term derived from the difference between the strong and weak signal:
\vspace{-0.2cm}
\begin{align}
    \mathbf{u}_{\text{w2s}} &= \mathbf{u} + w \cdot \mathbf{g}(\mathbf{x}_t,t,\mathbf{c})
\end{align}
This modification encourages the strong model to move beyond the conservative fit of standard MSE training and explcitely improve its extrapolative capacity. The training objective is:
\vspace{-0.2cm}
\begin{align}
\begin{split}
\mathcal{L}_s = \mathbb{E}_{t,\mathbf{x}_0,\boldsymbol{\epsilon}}\Big[&\big\|\mathbf{v}_{\theta}(\mathbf{x}_t, t, \mathbf{c}) \\ 
 & - \big( \mathbf{u} + w \cdot \text{sg}[\mathbf{g}(\mathbf{x}_t,t,\mathbf{c})] \big) \big\|_2^2 \Big]
\end{split}
\end{align}
Stop-gradient ($\text{sg}$) is used to stablize training, following protocols of~\cite{tang2025diffusion, chenvisual,gengmean}. The primary model network serves as the strong model. The main design choice lies in constructing an effective and efficient weak signal $\mathbf{v}_{\text{weak}}$. 

\noindent\textbf{Construction of Weak Signals for Training.}
We adapt existing inference-time guidance methods for the training phase and introduce a novel, highly efficient Condition-Agnostic Guidance (CAG) variant:

\begin{itemize}
    \item \textbf{CDG: CFG/MG.} Migrating the unconditional term ($\mathbf{v}(x_t, t, \emptyset)$) in CFG~\cite{ho2021classifier} into the training objective, an approach similar to MG~\cite{tang2025diffusion}.

    \item \textbf{CAG: AG.} Following AutoGuidance~\cite{karras2024guiding}, we maintain a separate, smaller and less-trained network during training to function as the weak model.

    \item \textbf{CAG: BR.} Inspired by the sequential structure of transformer blocks, this approach generates the weak signal by supervising an auxiliary output \textbf{br}anching from an intermediate layer.
\end{itemize}

BR is condition-agnostic and requires no extra forward calls during training for guidance. We also explored layer-perturbation methods (\eg, SLG~\cite{hyung2025spatiotemporal}) but found they degraded performance when integrated into training, thus excluded them (Further discussion are provided in the appendix).
Subsequently, we apply the idea of \textbf{Segmented Guidance (SGG)} directly to the training framework. The training-time version of SGG uses the condition-dependent guidance (CFG) signal for high noise levels ($t \ge \tau$) and switches to the condition-agnostic guidance (BR) signal for low noise levels ($t < \tau$). Illustration of the pipeline is provided in~\Cref{fig:arch}.

\section{Experiments}
\label{sec:experiments}

We validate our methods in two settings. First, we demonstrate the effectiveness of our inference-time SGG on state-of-the-art text-to-image models (SD3, SD3.5~\cite{esser2024scaling}). Second, to perform a controlled and computationally feasible analysis of our training-time integration, we follow standard practice~\cite{yurepresentation} and use the SiT-B/2 model on ImageNet, which allows us to ablate the W2S training targets (MG, AG, BR, and SGG) and measure the impact on training convergence and generalization.

\begin{table*}[htbp]
    \centering
    \setlength{\tabcolsep}{6.3pt}
    \renewcommand{\arraystretch}{1.1}
    
    \begin{tabular}{l|c|cccc|cccc}
    \toprule
     \multicolumn{1}{c|}{\textbf{Models}}& & \multicolumn{4}{c|}{SD3-medium} & \multicolumn{4}{c}{SD3.5-medium} \\
    \cmidrule(lr){3-6} \cmidrule(lr){7-10}
    
    \multicolumn{1}{c|}{\textbf{Dataset}} & & \multicolumn{2}{c|}{MS-COCO-1K} & \multicolumn{2}{c|}{LAION-5B-1K} & \multicolumn{2}{c|}{MS-COCO-1K} & \multicolumn{2}{c}{LAION-5B-1K} \\
    \cmidrule(lr){3-4} \cmidrule(lr){5-6} \cmidrule(lr){7-8} \cmidrule(lr){9-10}
    
    \multicolumn{1}{c|}{\textbf{Metrics}}& NFE/s& HPSv2.1 & Aes. & HPSv2.1 & Aes. & HPSv2.1 & Aes. & HPSv2.1 & Aes. \\
    \midrule
    
    Conditional (w/o Guidance) & 1 & 21.118 & 4.864 & 20.215 & 4.938 & 21.204 & 4.978 & 20.801 & 4.978 \\
    CFG~\cite{ho2021classifier} & 2 & 29.199 & 5.267 & 28.174 & 5.193 & 29.199 & 5.279 & 28.333 & 5.203 \\
    SLG~\cite{hyung2025spatiotemporal} & 2 & 26.685 & \underline{5.581} & 25.000 & 5.415 & 27.295 & \underline{5.714} & 26.050 & \underline{5.512} \\
    CFG+SLG~\cite{hyung2025spatiotemporal} & 3 & 27.759 & 5.468 & 26.147 & 5.285 & 28.331 & 5.678 & 27.246 & 5.421 \\
    \midrule
    CFG-zero-star~\cite{fan2025cfgzerostar} & 2 & 28.296 & 5.250 & 28.139 & 5.259 & 27.954 & 5.249 & 28.272 & 5.244 \\
    $S^2$ Guidance~\cite{chen2025s2guidancestochasticselfguidance} & 3 & \textbf{29.614} & 5.333 & \underline{28.442} & 5.244 & \underline{29.614} & 5.342 & \underline{28.491} & 5.250 \\
    Rectified CFG++~\cite{saini2025rectifiedcfgpp} & 3 & 28.932 & 5.399 & 28.306 & \underline{5.443} & 28.540 & 5.425 & 28.122 & 5.406 \\
    Guidance Interval~\cite{Kynkaanniemi2024} & 2 & 29.126 & 5.321 & 28.174 & 5.257 & 29.077 & 5.326 & 28.125 & 5.254 \\
    \midrule
    
    \rowcolor{blue!10}
    SGG & 2 & \underline{29.541} & \textbf{5.614} & \textbf{28.638} & \textbf{5.489} & \textbf{29.736} & \textbf{5.717} & \textbf{28.685} & \textbf{5.518} \\
    \bottomrule
    \end{tabular}
    
    \vspace{-0.1cm}
    \caption{Quantitative comparison of guidance methods on MS-COCO-1K and LAION-5B-1K, evaluating both HPSv2.1 and aesthetic scores for SD3 and SD3.5 models. Best results are in \textbf{bold}, second-best are \underline{underlined}.}
    \vspace{-0.5cm}
    \label{tab:guidance_comparison_sd3_sd35}
\end{table*}

\subsection{Implementation details}

\noindent\textbf{Inference-time guidance.}
For pre-trained model, we use the SD3-Medium and SD3.5-Medium as base models~\cite{esser2024scaling}. We use MS-COCO-1k~\cite{lin2014microsoft} subset and LAION-1k~\cite{schuhmann2022laion} subset for prompt instantiation.
We compare our method against several baselines, including standard conditional generation (no guidance), CFG~\cite{ho2021classifier}, and Skip-Layer Guidance (SLG). We also include comparisons to recent advanced guidance variants, such as $S^2$-Guidance~\cite{chen2025s2guidancestochasticselfguidance}, Guidance Interval~\cite{Kynkaanniemi2024}, CFG+SLG~\cite{hyung2025spatiotemporal}, CFG-Zero*~\cite{fan2025cfgzerostar} and Rectified-CFG++~\cite{saini2025rectifiedcfgpp}. 
We use the standard 28 inference steps throughout experiments. All methods are evaluated using HPSv2.1 Score~\cite{wu2023human} and Aesthetic Score~\cite{schuhmann2022aesthetics}. We select standard CFG~\cite{ho2021classifier} as CDG and SLG~\cite{hyung2025spatiotemporal} as CAG in SGG implementations.

\noindent\textbf{Training-time guidance}. We conduct training evaluation mainly on SiT-B/2 model~\cite{ma2024sit} due to computational constraints. We use lognormal-timestep sampling throughout all experiments to boost convergence, following~\cite{shin2025deeply}. We perform experiments in both unconditional and conditional settings. CAG methods are applied in both settings, whereas CDG method is naturally applied only in conditional training. For the conditional setting. All models are trained for 400k iterations. The sampling configuration is SDE Euler-Maruyama
 sampler with steps=250. 
We report the FID, sFID and Inception Score for all methods.

\noindent\textbf{NFE/s \& time/it}. We report the NFE per sampling step (NFE/s) during sampling. We also report wall-clock time per training iteration, normalized by the baseline configuration's time (time/it) to track the computation of guidance during training. Details of the implementation configurations across experiments could be referred in the appendix.

\vspace{-0.3cm}
\subsection{Inference time comparison}
\vspace{-0.1cm}

\begin{figure*}[tp]
    \centering
    \includegraphics[width=0.9\linewidth]{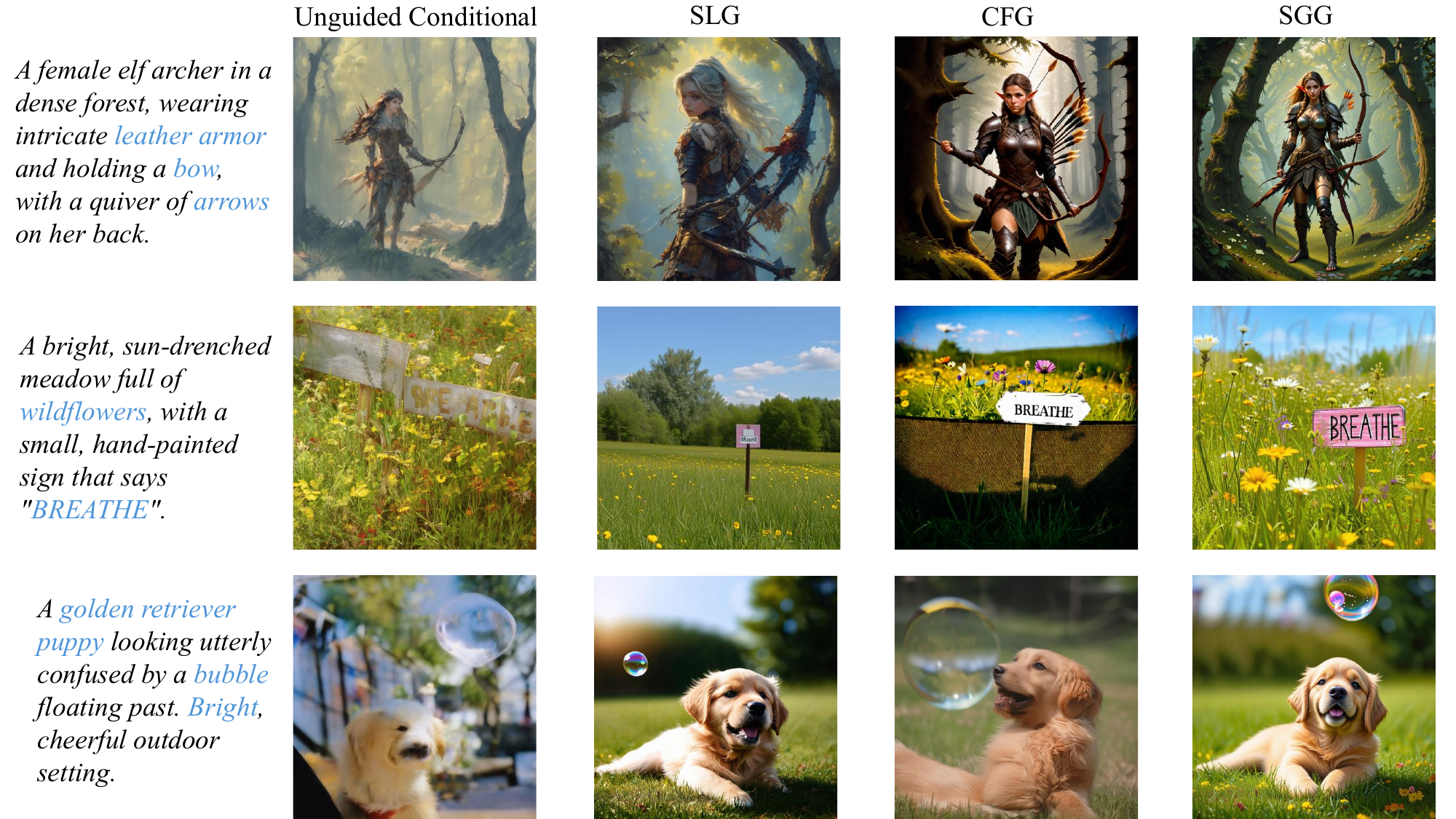}
    \caption{Qualitative comparison between Conditional (w/o guidance), CFG~\cite{ho2021classifier}, SLG~\cite{hyung2025spatiotemporal}, SGG (Ours).}
    \vspace{-0.5cm}
    \label{fig:qualitative}
\end{figure*}

We first conduct experiments to validate the effectiveness of our Segmented Guidance (SGG) principle against standard CFG and other prevalent guidance variants. 
As shown in~\Cref{tab:guidance_comparison_sd3_sd35}, a clear compromise between prompt-adherence (correlated to HPSv2.1) and aesthetic quality is evident in using CFG or SLG alone.
For example, on the SD3.5/MS-COCO benchmark, SLG achieves a high aesthetic score (5.714), but at a significant cost to its HPSv2.1 score (27.295).  Conversely, standard CFG achieves a competitive HPSv2.1 score (29.199) but produces a comparatively low aesthetic score (5.279).
As a hybrid approach to take the benefits of two, our Segmented Guidance (SGG) achieves the competitive scores in both categories (HPSv2.1: 29.736 and Aesthetic: 5.717). This pattern holds across models and datasets in our evaluation, where SGG reach comparable results to other guidance variants.
We also provide qualitative comparison of our methods, As illustrated in~\Cref{fig:qualitative}.



\begin{table}[b]
    \centering
    \vspace{-0.4cm}
    \setlength{\tabcolsep}{3.5pt}
    \renewcommand{\arraystretch}{1.0}
    \scriptsize 

    \resizebox{\columnwidth}{!}{%
    \begin{tabular}{lccccc}
    \toprule
    Model & NFE/s & time/it & FID $\downarrow$ & sFID $\downarrow$ & IS $\uparrow$ \\
    \midrule
    \multicolumn{6}{c}{\textit{Conditional Generation}} \\
    \midrule
    SiT-B/2 & 1 & 1.00 & 31.22 & 6.41 & 49.59 \\
    ~+ CFG & 2 & 1.00 & 6.02 & 5.47 & 183.83 \\
    \rowcolor{blue!10} AG & 1 & 1.27 & 13.96 & 88.36 & 4.68 \\
    \rowcolor{blue!10} BR & 1 & 1.02 & 16.02 & 5.13 & 76.21 \\
    \rowcolor{blue!10} MG & 1 & 1.23 & 5.88 & 6.19 & 253.74 \\
    \rowcolor{blue!10} SGG & 1 & 1.22 & \textbf{4.58} & \textbf{4.95} & \textbf{264.06} \\
    \midrule
    SiT-B/2+REPA & 1 & 1.00 & 20.46 & 6.31 & 73.00 \\
    \rowcolor{blue!10} SGG+REPA & 1 & 1.19 & \textbf{3.07} & \textbf{4.88} & \textbf{242.15} \\
    \midrule
    \multicolumn{6}{c}{\textit{Unconditional Generation}} \\
    \midrule
    SiT-B/2 & 1 & 1.00 & 61.27 & 7.00 & 17.33 \\
    \rowcolor{blue!10} AG & 1 & 1.26 & 45.97 & \textbf{4.94} & 20.32 \\
    \rowcolor{blue!10} BR & 1 & 1.02 & \textbf{43.25} & 5.11 & \textbf{20.66} \\
    \bottomrule
    \end{tabular}%
    }

    \caption{Training-time integration results on ImageNet 256$\times$256 with SiT-B/2, in conditional and unconditional settings.}
    \label{tab:training_results}
\end{table}

\subsection{Training convergence acceleration}

We subsequently evaluate the effectiveness of the migration of the weak-to-strong principle to boost training convergence, in both conditional and unconditional settings (\Cref{tab:training_results}).
In the conditional setting, our experiments demonstrate that all weak-to-strong guidance integrations consistently outperform the baseline.  with the hybrid SGG approach yields the best result. In unconditional setting, where CDG is not applicable, CAG methods (\eg BR, AG) still provide a notable performance boost over the baseline. We also observe that SGG in training time can be complemented with REPA~\cite{yurepresentation}, providing further performance gains.

Training integration of W2S guidance introduces a an extra forward call per iteration, \eg, an additional 22\% for the full SGG method. The cost could be offset by the resulting model's inference efficiency. The trained model's unguided output (NFE/s=1) achieves an FID of 4.58, which is superior to the guided (NFE/s=2) output of the baseline model (FID 6.02). Furthermore, BR variant in CAG incurs only a 2\% training overhead. This minimal cost still yields a reasonable FID improvement over the baseline, from 31.22 to 16.02 in conditional setting and from 61.27 to 43.25 in unconditional setting.


\begin{figure}[tp]
    \centering
    \includegraphics[width=0.98\linewidth]{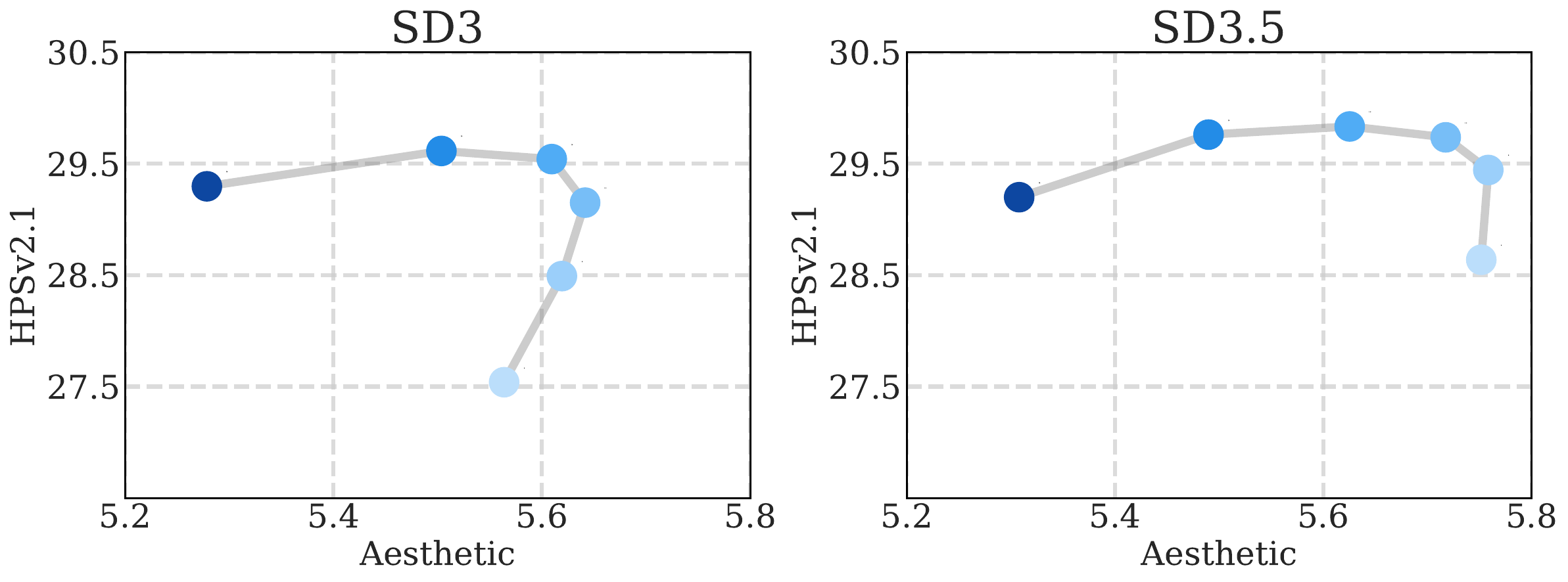}
    \caption{Ablation study on the segmentation timestep $\tau$. We vary $\tau$ from 4 (lightest) to 24 (darkest) in increments of 4, out of 28 total sampling steps. The results indicate that a mid-range segmentation point yields the best performance.}
    \vspace{-1.0cm}
    \label{fig:ablations}
\end{figure}

\begin{table}[b]
    \centering
    \vspace{-0.5cm}
    \resizebox{0.48\textwidth}{!}{
        \setlength{\tabcolsep}{8pt} \renewcommand{\arraystretch}{1.1}
    
    \
    \begin{tabular}{lccccc}
    \toprule
    Segmented \(\tau\) & 0.0 & 0.1 & 0.2 & 0.3 & 0.4 \\
    \midrule
FID $\downarrow$ & 5.88 & 5.26 & 4.58 & 4.99 & 5.79 \\
sFID $\downarrow$ & 6.19 & 5.44 & 4.95 & 5.37 & 6.37 \\
Inception Score $\uparrow$ & 253.74 & 267.34 & 264.06 & 249.52 & 236.51 \\
\bottomrule
    \end{tabular}
    }
    \caption{Ablation study on segmentation timestamp \(\tau\) conditional training with SGG on ImageNet 256x256. We choose \(\tau=0.2\) as our default setting.}
    \label{tab:ablation_tau}
    \vspace{-1.0cm}
\end{table}

\vspace{-0.2cm}
\subsection{Ablation study}
\vspace{-0.1cm}
We ablate two critical components: (1) The segmented timestep \(\tau\) between CDG and CAG. (2) The guidance weight \(w\).
As illustrated in~\Cref{fig:ablations}, our ablation on the guidance segmentation point reveals a Pareto frontier.  This frontier traces the trade-off between HPSv2.1 (prompt adherence) and aesthetic score as the segmented step transitions from high noise level to low noise level. We also conducted ablation on \(\tau\) in conditional training configuration, shown in~\Cref{tab:ablation_tau}. 

\vspace{-0.3cm}
\section{Conclusion}
\label{sec:conclusion}
\vspace{-0.2cm}
In this work, we first clarify the generalization issues in common diffusion models and the alleviation by guidance. We then systematically analyze the operational regimes of condition-dependent and condition-agnostic approaches under the perspective of weak-to-strong principle. 
Based on this analysis, we proposed Segmented Guidance (SGG), a simple and effective approach that synergizes the benefits of both guidance types. We subsequently migrate W2S principle along with SGG into the training objective, thereby reducing the need for guidance during inference. Comprehensive qualitative and quantitative comparisons validate the effectiveness of both Segmented Guidance and training-time integration of weak-to-strong principle.

\noindent\textbf{Limitations and future work.}
Our approach is limited to continuous diffusion,
future work could benefit from migrating the segmentation idea of SGG to other modalities (\eg discrete diffusion) and further explore the combination of different guidance instances under W2S principle. 

\section*{Acknowledgement}

This work was supported by the National Natural Science Foundation of China (No. 6250070674) and the Zhejiang Leading Innovative and Entrepreneur Team Introduction Program (2024R01007).

{\small
\bibliographystyle{ieeenat_fullname}
\bibliography{sections/11_references}
}

\ifarxiv \clearpage \appendix \ifarxiv
\else
  \begin{document}
\fi

\ifarxiv
\else
  \title{\paperTitle}
  \author{\authorBlock}
  \maketitlesupplementary
\fi

\appendix

\tableofcontents

\section{Error correction analysis on ImageNet}
\label{sec:error_correction_experiments_on_imagenet}

\subsection{Derivation of optimal conditional velocity}

Previous work has derived the optimal denoiser $\mathbf{D}(\mathbf{x}_t,t)$~\cite{Karras2022edm} and score-matching objective $\mathbf{s}(\mathbf{x}_t,t)$~\cite{gumemorization}. Here, we provide a detailed derivation of the optimal conditional velocity, $\dot{\mathbf{v}}(\mathbf{x}_t, t, \mathbf{\mathbf{c}})$, given a state $(\mathbf{x}_t, t)$ and a condition $\mathbf{\mathbf{c}}$.

We adopt the flow matching (OT)~\cite{lipmanflow,liuflow,albergo2023stochastic} schedule, where $\alpha_t = 1-t$, $\sigma_t = t$, and the state is $\mathbf{x}_t = (1-t)\mathbf{x}_0 + t\epsilon$. The corresponding true velocity is $\mathbf{u} = \epsilon - \mathbf{x}_0$.

The optimal conditional velocity $\dot{\mathbf{v}}(\mathbf{x}_t, t, \mathbf{\mathbf{c}})$ is the function that minimizes the mean-squared error. This is achieved by the conditional expectation of the true velocity, given the current state and condition:
\begin{equation}
    \dot{\mathbf{v}}(\mathbf{x}_t, t, \mathbf{\mathbf{c}}) = \mathbb{E}_{\mathbf{x}_0 \sim p(\cdot|\mathbf{\mathbf{c}}), \epsilon}[\epsilon - \mathbf{x}_0 \mid \mathbf{x}_t, t, \mathbf{\mathbf{c}}]
\end{equation}
We can simplify this expression. Given $\epsilon = (\mathbf{x}_t - (1-t)\mathbf{x}_0) / t$, the true velocity $\mathbf{u}$ becomes:
\begin{equation}
    \mathbf{u} = \epsilon - \mathbf{x}_0 = \frac{\mathbf{x}_t - (1-t)\mathbf{x}_0}{t} - \mathbf{x}_0 = \frac{\mathbf{x}_t - \mathbf{x}_0}{t}
\end{equation}
Substituting this back into the expectation, and noting that $\mathbf{x}_t$ and $t$ are given:
\begin{align}
    \dot{\mathbf{v}}(\mathbf{x}_t, t, \mathbf{\mathbf{c}}) &= \mathbb{E}_{\mathbf{x}_0 \sim p(\cdot|\mathbf{\mathbf{c}})}\left[ \frac{\mathbf{x}_t - \mathbf{x}_0}{t} \mid \mathbf{x}_t, t, \mathbf{\mathbf{c}} \right] \\
    &= \frac{1}{t} \left( \mathbf{x}_t - \mathbb{E}[\mathbf{x}_0 \mid \mathbf{x}_t, t, \mathbf{\mathbf{c}}] \right) \label{eq:vel_to_posterior}
\end{align}
The problem thus reduces to finding the posterior mean $\mathbb{E}[\mathbf{x}_0 \mid \mathbf{x}_t, t, \mathbf{\mathbf{c}}]$. We find the posterior distribution $p(\mathbf{x}_0 \mid \mathbf{x}_t, t, \mathbf{\mathbf{c}})$ using Bayes' rule. Note that the perturbation kernel $p_{0t}$ is independent of $\mathbf{\mathbf{c}}$:
\begin{equation}
    p(\mathbf{x}_0 \mid \mathbf{x}_t, t, \mathbf{\mathbf{c}}) = \frac{p_{0t}(\mathbf{x}_t \mid \mathbf{x}_0) p(\mathbf{x}_0 \mid \mathbf{\mathbf{c}})}{p_t(\mathbf{x}_t \mid \mathbf{\mathbf{c}})}
\end{equation}
We now assume a finite dataset. Let the subset of data points belonging to condition $\mathbf{\mathbf{c}}$ be a finite set of $N$ samples, $\{\mathbf{x}_0^i\}_{i=1}^N$. The conditional data distribution $p(\mathbf{x}_0 \mid \mathbf{\mathbf{c}})$ can be expressed as a sum of Dirac delta functions:
\begin{equation}
    p(\mathbf{x}_0 \mid \mathbf{\mathbf{c}}) = \frac{1}{N}\sum_{i=1}^N\delta(\mathbf{x}_0 - \mathbf{x}_0^i)
\end{equation}
The denominator, $p_t(\mathbf{x}_t \mid \mathbf{\mathbf{c}})$, is the conditional marginal probability:
\begin{align}
    &p_t(\mathbf{x}_t \mid \mathbf{\mathbf{c}}) = \int p_{0t}(\mathbf{x}_t \mid \mathbf{x}_0) p(\mathbf{x}_0 \mid \mathbf{\mathbf{c}}) \mathrm{d}\mathbf{x}_0 \\
    &= \int \mathcal{N}(\mathbf{x}_t;(1-t)\mathbf{x}_0,t^2\mathbf{I}) \left(\frac{1}{N}\sum_{i=1}^N\delta(\mathbf{x}_0 - \mathbf{x}_0^i)\right) \mathrm{d}\mathbf{x}_0 \\
    &= \frac{1}{N}\sum_{i=1}^N \mathcal{N}(\mathbf{x}_t; (1-t)\mathbf{x}_0^i, t^2\mathbf{I})
\end{align}
With the numerator and denominator defined, the full posterior $p(\mathbf{x}_0 \mid \mathbf{x}_t, t, \mathbf{\mathbf{c}})$ is a weighted sum of Dirac deltas:
\begin{equation}
    p(\mathbf{x}_0 \mid \mathbf{x}_t, t, \mathbf{\mathbf{c}}) = \frac{\sum_{i=1}^N \mathcal{N}(\mathbf{x}_t; (1-t)\mathbf{x}_0^i, t^2\mathbf{I}) \delta(\mathbf{x}_0 - \mathbf{x}_0^i)}{\sum_{j=1}^N \mathcal{N}(\mathbf{x}_t; (1-t)\mathbf{x}_0^j, t^2\mathbf{I})}
\end{equation}
The posterior mean $\mathbb{E}[\mathbf{x}_0 \mid \mathbf{x}_t, t, \mathbf{\mathbf{c}}]$ is therefore the weighted average of the conditional data points $\{\mathbf{x}_0^i\}$:
\begin{align}
    \mathbb{E}[\mathbf{x}_0 \mid \mathbf{x}_t, t, \mathbf{\mathbf{c}}] &= \int \mathbf{x}_0 p(\mathbf{x}_0 \mid \mathbf{x}_t, t, \mathbf{\mathbf{c}}) \mathrm{d}\mathbf{x}_0 \\
    &= \frac{\sum_{i=1}^N \mathbf{x}_0^i \mathcal{N}(\mathbf{x}_t; (1-t)\mathbf{x}_0^i, t^2\mathbf{I})}{\sum_{j=1}^N \mathcal{N}(\mathbf{x}_t; (1-t)\mathbf{x}_0^j, t^2\mathbf{I})} \label{eq:posterior_mean}
\end{align}
Finally, we substitute this posterior mean (Eq. \ref{eq:posterior_mean}) back into our velocity expression (Eq. \ref{eq:vel_to_posterior}):
\begin{align}
    \dot{\mathbf{v}}(\mathbf{x}_t, t, \mathbf{\mathbf{c}}) &= \frac{1}{t} \left( \mathbf{x}_t - \frac{\sum_{i=1}^N \mathbf{x}_0^i \mathcal{N}(\cdot)}{\sum_{j=1}^N \mathcal{N}(\cdot)} \right) \\
    &= \frac{1}{t} \left( \frac{\mathbf{x}_t \sum_{j=1}^N \mathcal{N}(\cdot) - \sum_{i=1}^N \mathbf{x}_0^i \mathcal{N}(\cdot)}{\sum_{j=1}^N \mathcal{N}(\cdot)} \right) \\
    &= \frac{\sum_{i=1}^N (\mathbf{x}_t - \mathbf{x}_0^i) \mathcal{N}(\mathbf{x}_t; (1-t)\mathbf{x}_0^i, t^2\mathbf{I})}{t \sum_{j=1}^N \mathcal{N}(\mathbf{x}_t; (1-t)\mathbf{x}_0^j, t^2\mathbf{I})}
\end{align}
Given the set of data points $\{\mathbf{x}_0^i\}_{i=1}^N$ corresponding to condition $\mathbf{\mathbf{c}}$, this equation provides the exact velocity target.

\subsection{Experiments configurations}
\label{subsec:experimetns_configuration}

\noindent\textbf{Pretraind model configuration}. 
For the comparison of CFG and AG against optimal velocity in~\Cref{fig:lossvsguidance}, we pre-trained a SiT-B/2 model for 400k iterations to serve as the strong model. We also pre-trained a SiT-S/2 model for 100k iterations to serve as the weak model required for AutoGuidance (AG)~\cite{karras2024guiding}.

\noindent\textbf{Inception distance between guided velocity and optimal velocity}. 
For the inception distance~\cite{heusel2017gans} analysis in~\Cref{fig:lossvsguidance}, we use the standard extrapolation guidance formula $\mathbf{v}_{w} = \mathbf{v}_{\text{cond}} + w \cdot (\mathbf{v}_{\text{cond}} - \mathbf{v}_{\text{weak}})$.
For CFG~\cite{ho2021classifier}, we use unconditional output \(\mathbf{v}(\mathbf{x}_t,t,\emptyset)\) as \(\mathbf{v}_{\text{weak}}\). For AG~\cite{karras2024guiding}, we use the condition-aligned output \(\tilde{\mathbf{v}}(\mathbf{x}_t,t,\mathbf{\mathbf{c}})\) from SiT-S/2 as \(\mathbf{v}_{\text{weak}}\).
We tested extrapolation scales of $w \in \{1.0, 1.2, 1.4, 1.6\}$.  
The $w=1.0$ case represents the unguided conditional output ($\mathbf{v}_{w} = \mathbf{v}_{\text{cond}}$) and serves as our baseline for comparison. For a given class \(\mathbf{\mathbf{c}}\), we calculate the following distance objective:

\begin{equation}
\mathbb{E}_{t,\mathbf{x}_t}\|\dot{\mathbf{v}}(\mathbf{x}_t,t,\mathbf{\mathbf{c}})-\mathbf{v}_{w}(\mathbf{x}_t,t,\mathbf{\mathbf{c}})\|_2^2
\end{equation}  

We sample 100000 samples to calculate the corresponding state \((\mathbf{x}_t,t)\) given timestamp \(t\).

\section{Toy experiment implementation}
\label{sec:toy-example-details}

\subsection{Network architectures}
\label{sec:toy-arch}

For all 2D toy experiments we train a class-conditional diffusion model with a velocity
parameterization
\begin{equation}
    \mathbf{v} : \mathbb{R}^2 \times [0,1] \times \mathcal{C} \to \mathbb{R}^2,
    \qquad
    (\mathbf{x}_t, t, \mathbf{c}) \mapsto \mathbf{v}(\mathbf{x}_t, t, \mathbf{c}),
\end{equation}
where $\mathbf{x}_t \in \mathbb{R}^2$ is the noisy state, $t \in [0,1]$ is the continuous time index, and
$\mathbf{c} \in \mathcal{C} = \{1,\dots,\mathrm{CLS}\}$ is the class label.

For network architectural design, we follow the score network in~\cite{karras2024guiding} but apply the following modification to enable conditional and unconditional prediction on the same model architecture. We introduce a learnable class-embedding table
\begin{equation}
    E : \mathcal{C} \cup \{\emptyset\} \to \mathbb{R}^{d_c},
    \qquad
    \mathbf{e}_c = E(\mathbf{c}),
\end{equation}
where $\mathbf{c} = \emptyset$ denotes the unconditional (null) condition and $d_c$ is the embedding dimension. Let $\mathrm{Enc}(\mathbf{x}_t, t) \in \mathbb{R}^{d_h}$ denote the standard feature encoding of the noisy state and time as in~\cite{karras2024guiding}. We then form the input to the backbone as
\begin{equation}
    \mathbf{h}_0
    =
    \big[
        \mathrm{Enc}(\mathbf{x}_t, t)
        \,;\,
        \mathbf{e}_c
    \big]
    \in \mathbb{R}^{d_h + d_c},
\end{equation}
and use the same network weights for all $\mathbf{c} \in \mathcal{C} \cup \{\emptyset\}$. This design allows us to obtain both
\begin{equation}
    \mathbf{v}(\mathbf{x}_t, t, \mathbf{c})
    \quad\text{and}\quad
    \mathbf{v}(\mathbf{x}_t, t, \emptyset)
\end{equation}
from a single model, thereby enabling classifier-free guidance~\cite{ho2021classifier} and our segmented guidance without changing the backbone.

Training follows the flow-matching parameterization adopted in the main paper. We sample $\mathbf{x}_0 \sim p_{\mathrm{data}}(\mathbf{x}_0 \mid \mathbf{c})$ and $\epsilon \sim \mathcal{N}(\mathbf{0}, \mathbf{I}_2)$, and construct the noisy state via the stochastic interpolant
\begin{equation}
    \mathbf{x}_t = (1 - t)\,\mathbf{x}_0 + t\,\epsilon,
    \qquad
    t \sim p(t),
\end{equation}
where $p(t)$ is the lognormal time sampling distribution used throughout the paper. The network is trained with the standard velocity regression objective
\begin{equation}
    \mathcal{L}_{\mathrm{toy}}(\theta)
    =
    \mathbb{E}_{t, \mathbf{c}, \mathbf{x}_0, \epsilon}
    \big[
        \big\|
            \mathbf{v}(\mathbf{x}_t, t, \mathbf{c})
            -
            (\epsilon - \mathbf{x}_0)
        \big\|_2^2
    \big].
\end{equation}

\subsection{Construction of the toy dataset}
\label{sec:toy-data}

Our toy dataset construction shares the principle of using a mixture of Gaussians as the building block as in~\cite{karras2024guiding}, but explicitly exposes the \emph{granularity of the condition} via the number of classes and the recursive depth. The dataset returns a Gaussian mixture distribution constructed from a collection of leaf- and branch-like components.

We denote the class set by
\begin{equation}
    \mathcal{C} = \{1,\dots,\mathrm{CLS}\},
\end{equation}
corresponding to the \texttt{num\_classes} argument. Internally, the function assembles a list of Gaussian components indexed by $i = 1,\dots,K$, each with a weight $\phi_i > 0$, mean $\boldsymbol{\mu}_i \in \mathbb{R}^2$, covariance $\boldsymbol{\Sigma}_i \in \mathbb{R}^{2 \times 2}$, and a discrete class label $c_i \in \mathcal{C} \cup \{c_{\mathrm{base}}\}$. After selecting a subset of labels through the \texttt{classes} argument, the final distribution is the normalized Gaussian mixture
\begin{equation}
    p_{\mathrm{data}}(\mathbf{x}_0, \mathbf{c})
    =
    p(\mathbf{c})\,p_{\mathrm{data}}(\mathbf{x}_0 \mid \mathbf{c}),
    \qquad
    p(\mathbf{c}) = \frac{1}{|\mathcal{C}_{\mathrm{sel}}|},
\end{equation}
where $\mathcal{C}_{\mathrm{sel}} \subseteq \mathcal{C}$ is the set of selected class labels and
\begin{equation}
    p_{\mathrm{data}}(\mathbf{x}_0 \mid \mathbf{c})
    =
    \sum_{i \in I_c}
    \pi_i^{(\mathbf{c})}\,
    \mathcal{N}\!\big(
        \mathbf{x}_0;\,
        \boldsymbol{\mu}_i,\,
        \boldsymbol{\Sigma}_i
    \big),
    \qquad
    \sum_{i \in I_c} \pi_i^{(\mathbf{c})} = 1.
\end{equation}
Here $I_c = \{ i : c_i = \mathbf{c} \}$ collects all components assigned to class $\mathbf{c}$, and the mixture weights $\pi_i^{(\mathbf{c})}$ are obtained by normalizing the raw branch weights $\phi_i$ within the class.

The geometry of the mixture components is determined by a recursive branching construction. A single ``main branch'' of length
\begin{equation}
    L_{\mathrm{main}} = 0.4\,\bigl(1 + 0.1 \cdot \texttt{num\_classes}\bigr)
\end{equation}
is grown from a base point $\mathbf{x}_{\mathrm{base}} \in \mathbb{R}^2$ with initial angle $\alpha_{\mathrm{main}} \approx 85^\circ$. This branch is split into $\texttt{num\_classes}$ segments, and each segment serves as the attachment point for a class-specific subbranch.

Each subbranch is generated by a recursive procedure with maximum depth, branching factor, and curvature. At recursion depth $d \in \{0,\dots,\texttt{max\_depth}-1\}$, a subbranch located at position $\mathbf{p}^{(d)}$ with direction $\mathbf{u}^{(d)} \in \mathbb{R}^2$ and overall size $s^{(d)}$ generates a sequence of Gaussian components
\begin{equation}
    \boldsymbol{\mu}_{i}
    =
    \bigl(
        \mathbf{p}^{(d)}
        +
        \lambda\,\mathbf{u}^{(d)}
    \bigr)
    \odot
    \mathbf{s},
    \qquad
    \boldsymbol{\Sigma}_{i}
    =
    \mathbf{R}^{(d)}
    \,\mathbf{D}^{(d)}\,
    \mathbf{R}^{(d)\top},
\end{equation}
for several values of $\lambda \in (0,1)$ along the branch; here $\mathbf{s} = \texttt{scale} \in \mathbb{R}^2$ scales the coordinates, $\mathbf{R}^{(d)}$ is the $2\times 2$ rotation induced by the current branch angle, and $\mathbf{D}^{(d)}$ is a diagonal matrix encoding the anisotropic thickness of the branch. The raw weight of each component is proportional to a depth-dependent factor $\phi_i \propto s^{(d)} (0.6)^d$, which causes branch segments closer to the root to receive higher total mass.

\subsection{Guidance baselines and configurations}
\label{sec:toy-config}

We evaluate four guidance configurations on the above toy datasets: unguided sampling, condition-dependent guidance (CDG, instantiated by CFG~\cite{ho2021classifier}), condition-agnostic guidance (CAG, instantiated by AG~\cite{karras2024guiding}), and our proposed segmented guidance (SGG). All methods act on the same strong model $\mathbf{v}(\mathbf{x}_t, t, \mathbf{c})$, and differs only in how they construct the weak signal $\mathbf{v}_{\mathrm{weak}}$ within the weak-to-strong extrapolation.

\noindent\textbf{Unguided.}
We use the strong model directly,
\begin{equation}
    \mathbf{v}^{\mathrm{ung}}(\mathbf{x}_t, t, \mathbf{c})
    =
    \mathbf{v}(\mathbf{x}_t, t, \mathbf{c}),
\end{equation}
corresponding to $w = 1$.

\noindent\textbf{CDG: CFG}.
Here the weak signal is the unconditional prediction
$\mathbf{v}_{\mathrm{weak}}(\mathbf{x}_t, t, \mathbf{c})
=
\mathbf{v}(\mathbf{x}_t, t, \emptyset)$
and the strong signal is the class-conditional prediction
$\mathbf{v}_{\mathrm{strong}}(\mathbf{x}_t, t, \mathbf{c})
=
\mathbf{v}(\mathbf{x}_t, t, \mathbf{c})$.
This yields the usual classifier-free guidance form
\begin{align}
\begin{split}
    \mathbf{v}^{\mathrm{CFG}}_w(\mathbf{x}_t, t, \mathbf{c})
    &=
    \mathbf{v}(\mathbf{x}_t, t, \mathbf{c})
    +\\
    &(w - 1)
    \big(
        \mathbf{v}(\mathbf{x}_t, t, \mathbf{c})
        -
        \mathbf{v}(\mathbf{x}_t, t, \emptyset)
    \big).
\end{split}
\end{align}

\noindent\textbf{CAG: AG}
Following autoguidance~\cite{karras2024guiding}, we construct a weaker but condition-aligned model $\tilde{\mathbf{v}}_\theta(\mathbf{x}_t, t, \mathbf{c})$ by reducing capacity or early stopping. In this case the weak signal is
$\mathbf{v}_{\mathrm{weak}}(\mathbf{x}_t, t, \mathbf{c})
=
\tilde{\mathbf{v}}_\theta(\mathbf{x}_t, t, \mathbf{c})$,
which leads to
\begin{align}
\begin{split}
    \mathbf{v}^{\mathrm{AG}}_w(\mathbf{x}_t, t, \mathbf{c})
    &=
    \mathbf{v}(\mathbf{x}_t, t, \mathbf{c})
    + \\
    &(w - 1)
    \big(
        \mathbf{v}(\mathbf{x}_t, t, \mathbf{c})
        -
        \tilde{\mathbf{v}}(\mathbf{x}_t, t, \mathbf{c})
    \big).
\end{split}
\end{align}

\noindent\textbf{Segmented guidance (SGG, ours).}
SGG uses a time-dependent segmentation between CDG and CAG. For a switching time $\tau \in (0,1)$ we define
\begin{equation}
    \mathbf{g}(\mathbf{x}_t, t, \mathbf{c})
    =
    \begin{cases}
        \mathbf{v}(\mathbf{x}_t, t, \mathbf{c})
        -
        \mathbf{v}(\mathbf{x}_t, t, \emptyset),
        & t \ge \tau,\\[2pt]
        \mathbf{v}(\mathbf{x}_t, t, \mathbf{c})
        -
        \tilde{\mathbf{v}}_\theta(\mathbf{x}_t, t, \mathbf{c}),
        & t < \tau,
    \end{cases}
\end{equation}
and set
\begin{equation}
    \mathbf{v}^{\mathrm{SGG}}_w(\mathbf{x}_t, t, \mathbf{c})
    =
    \mathbf{v}(\mathbf{x}_t, t, \mathbf{c})
    +
    (w - 1)\,\mathbf{g}(\mathbf{x}_t, t, \mathbf{c}).
\end{equation}

To simulate the interplay between condition granularity and model fitting capacity, we vary the tuple
\[
(\text{CLS}, \text{Depth}, B)
\]
where $\text{CLS}$ is the number of classes, $\text{Depth}$ is the maximum recursion depth and $B$ is the number of branches per split, together with the training budget $T$ of the strong model. We consider three representative configurations:

\begin{itemize}
    \item \textbf{Config A (blurry condition, complex in-class).}
    This regime uses a small number of classes but a deep recursive structure,
    \[
        \text{CLS} = 4,\quad \text{Depth} = 3, \quad B = 2, \quad T = 2^{15},
    \]
    which yields \emph{blurry} conditions and highly intricate within-class manifolds (well-fitted but hard to disambiguate at the label level).

    \item \textbf{Config B (sharp condition, simple in-class).}
    This regime uses many classes but a shallow recursive structure,
    \[
        \text{CLS} = 24,\quad \text{Depth} = 1,\quad B = 2,\quad T = 2^{12},
    \]
    leading to \emph{sharp} conditions with relatively simple manifolds that are harder to fit under the limited training budget.

    \item \textbf{Config C (intermediate, realistic regime).}
    This regime interpolates between the above two,
    \[
        \text{CLS} = 12,\quad \text{Depth} = 2,\quad B = 2,\quad T = 2^{15},
    \]
    producing moderately complex intra-class structure together with non-trivial conditioning. The difficulty of this task is relatively higher than Config A and B, attempting to project to realistic scenarios. 
\end{itemize}

\noindent\textbf{Configurations across settings}. Here we provide the full setting on toy experiments across Configurations and hyperparameters, as illustrated in~\Cref{tab:toy-guidance-configs}.

\begin{table}[ht]
  \centering
  \footnotesize
  \setlength{\tabcolsep}{3.0pt}
  \renewcommand{\arraystretch}{1.00}
  \begin{tabular*}{\columnwidth}{@{\extracolsep{\fill}}lccccc}
    \toprule
    \textbf{Setting} & $T_{\mathrm{main}}$ & $T_{\mathrm{weak}}$ & $\tau$ & $w_{\mathrm{cfg}}$ & $w_{\mathrm{ag}}$ \\
    \midrule
    Config A & $2^{15}$ & $2^{11}$ & 0.5 & 2.0 & 2.0 \\
    Config B & $2^{12}$ & $2^{10}$ & 0.1 & 2.0 & 2.0 \\
    Config C & $2^{15}$ & $2^{11}$ & 0.3 & 2.0 & 2.0 \\
    \bottomrule
  \end{tabular*}
  \vspace{-0.5em}
  \caption{Hyperparameter settings for toy experimentation.}
  \label{tab:toy-guidance-configs}
\end{table}

\noindent\textbf{Limitations across dimensionality.}
Toy examples are powerful for visualizing algorithmic behavior at the distribution level, rather than relying solely on aggregate quantitative metrics~\cite{galashov2025learn,karras2024guiding,chen2025s2guidancestochasticselfguidance,gao2025reg}. 
However, there is an inherent gap between 2D toy class-conditional tasks, image class-conditional tasks, and image prompt-conditional (text-to-image) tasks, and phenomena observed in the 2D plane are not guaranteed to transfer to high-dimensional image space with \(100\%\) accuracy~\cite{FuTCFG2025,song2025selective}.
Our inductive bias in this work is to isolate the interplay between \emph{condition granularity} and \emph{fitting capacity} as the factors influencing the behavior of CDG and CAG. The 2D toy results should therefore be viewed as qualitative insight into these mechanisms, providing explanations for real image-generation setups.


\begin{figure*}[t]
    \centering
    \begin{subfigure}[t]{0.48\linewidth}
        \centering
        \includegraphics[width=\linewidth]{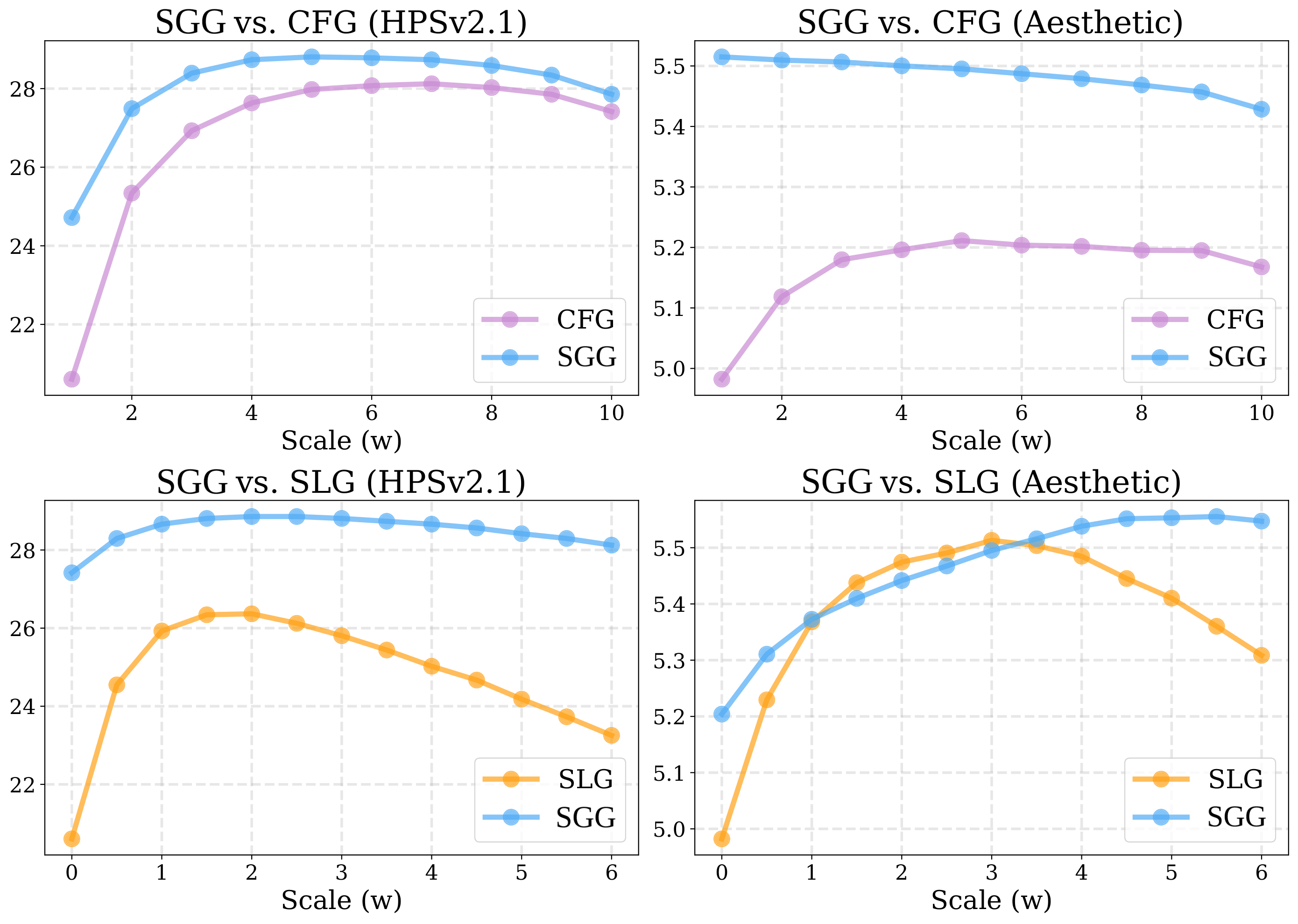}
        \caption{LAION-5B-1K}
        \label{fig:supp_ablation_grid_laion}
    \end{subfigure}
    \hfill 
    \begin{subfigure}[t]{0.48\linewidth}
        \centering
        \includegraphics[width=\linewidth]{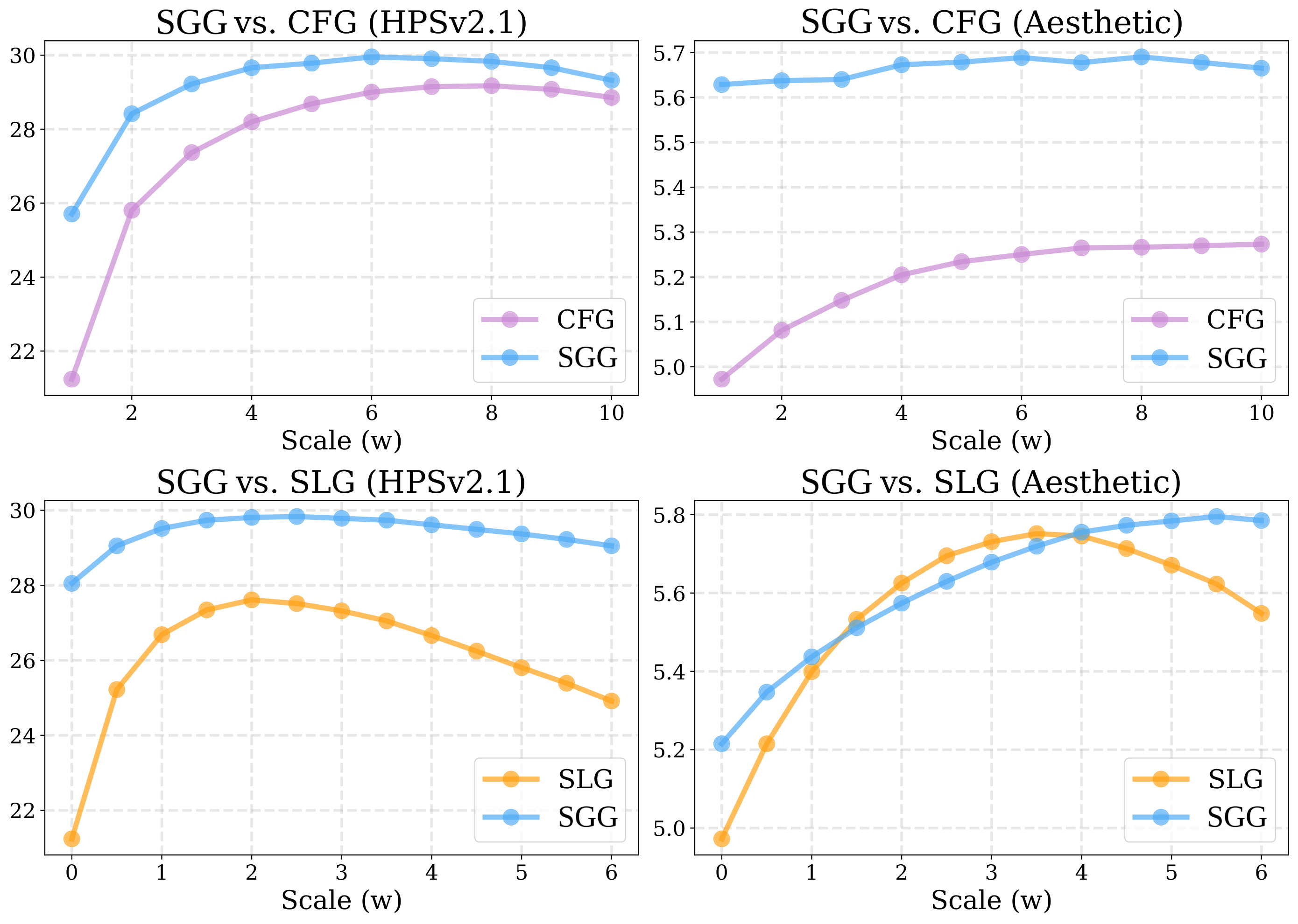}
        \caption{MSCOCO-1K}
        \label{fig:supp_ablation_grid_mscoco}
    \end{subfigure}
    
    \caption{Quantitative comparison of guidance scale ($w$) for CFG and SLG with SD3.5 on LAION-5B-1K and MSCOCO-1K datasets.}
    \label{fig:supp_ablation_grids}
\end{figure*}

\begin{table*}[b]
    \centering
    \resizebox{0.9\textwidth}{!}{%
    \renewcommand{\arraystretch}{1.1}
    \setlength{\tabcolsep}{6pt}
    \begin{tabular}{l|ccccccc}
    \toprule
    Method & Avg & Aesthetic & Overall Cons. & Imaging Qual. & Subject Cons. & Dynamic Deg. & Motion Smooth. \\
    \midrule
    CFG & \underline{0.6877} & \underline{0.6003} & \textbf{0.2317} & 0.6553 & 0.9391 & \underline{0.7222} & 0.9776 \\
    SLG & 0.6809 & 0.5822 & 0.2026 & \underline{0.6597} & \textbf{0.9504} & 0.7083 & \textbf{0.9820} \\
    \textbf{SGG} & \textbf{0.7001} & \textbf{0.6107} & \underline{0.2314} & \textbf{0.6624} & \underline{0.9402} & \textbf{0.7778} & \underline{0.9785} \\
    \bottomrule
    \end{tabular}%
    }
    \caption{Comparison on WAN-1.3B~\cite{wan2025wan} with CFG, SLG and SGG (Ours). Best results are \textbf{bolded}, and second-best results are \underline{underlined}.}
    \label{tab:supp_wan}
\end{table*}

\section{Inference implementation and ablations}

\subsection{Inference time settings}

\noindent\textbf{Pretrained models and baseline methods selection}. For pre-trained model, we use the SD3-Medium and SD3.5-Medium as base models~\cite{esser2024scaling}. We use MS-COCO-1k~\cite{lin2014microsoft} subset and LAION-1k~\cite{schuhmann2022laion} randomly selected subset for prompt instantiation.
We compare our method against several baselines, including standard conditional generation (no guidance), CFG~\cite{ho2021classifier}, and Skip-Layer Guidance (SLG). We also include comparisons to recent advanced guidance variants, such as $S^2$-Guidance~\cite{chen2025s2guidancestochasticselfguidance}, Guidance Interval~\cite{Kynkaanniemi2024}, CFG+SLG~\cite{hyung2025spatiotemporal}, CFG-Zero*~\cite{fan2025cfgzerostar} and Rectified-CFG++~\cite{saini2025rectifiedcfgpp}. 
We use the standard 28 inference steps throughout experiments. All methods are evaluated using HPSv2.1 Score~\cite{wu2023human} and Aesthetic Score~\cite{schuhmann2022aesthetics}. We select standard CFG~\cite{ho2021classifier} as CDG and SLG~\cite{hyung2025spatiotemporal} as CAG in SGG implementations. 

\noindent\textbf{Hyperparameter settings.}
For standard CFG~\cite{ho2021classifier}, we performed a grid search for the guidance scale $w$ in the range $[1.0, 9.0]$ with an interval of 0.5, selecting the optimal value of $w=5$. For Guidance Interval~\cite{Kynkaanniemi2024}, we searched for the optimal interval $t$ with a step of 0.1, finding that removing guidance for the $20\%$ of timestamps closest to the data ($t < 0.2$) yielded the best results. For $S^2$-Guidance~\cite{chen2025s2guidancestochasticselfguidance}, CFG+SLG~\cite{hyung2025spatiotemporal}, CFG-Zero*~\cite{fan2025cfgzerostar}, and Rectified-CFG++~\cite{saini2025rectifiedcfgpp}, we adhered to the recommended hyperparameter settings from their respective papers. For our method, SGG, we set the segmentation timestamp $\tau=12/28~(t_m = 0.69,\text{SD3.5}),~~\tau=16/28~(t_m = 0.8, \text{SD3})$ and use a scale of $w=5$ for the CDG (CFG) component and $w=3$ for the CAG (SLG) component for both models. The skipping layers are the default setting in vanilla SLG with 7,8,9.

\subsection{Ablations on inference guidance scale}

\begin{table*}[b]
    \centering
    \resizebox{0.98\textwidth}{!}{%
    \setlength{\tabcolsep}{9pt}
    \renewcommand{\arraystretch}{1.1}
    
    \begin{tabular}{lcccccccc}
    \toprule
    & \multicolumn{5}{c}{Conditional} & \multicolumn{3}{c}{Unconditional} \\
    \cmidrule(lr){2-6} \cmidrule(lr){7-9}
    Parameter & Baseline & AG & BR & MG & SGG & Baseline & AG & BR \\
    \midrule
    
    \textbf{Training configuration} & & & & & & & & \\
    \quad Generation Type & Cond & Cond & Cond & Cond & Cond & Uncond & Uncond & Uncond \\
    \quad Batch-size & 256 & 256 & 256 & 256 & 256 & 256 & 256 & 256 \\
    \quad Num. GPUs (A100) & 4 & 4 & 4 & 4 & 4 & 4 & 4 & 4 \\
    \quad Noise Schedules & OT & OT & OT & OT & OT & OT & OT & OT \\
    \quad LR & 0.0001 & 0.0001 & 0.0001 & 0.0001 & 0.0001 & 0.0001 & 0.0001 & 0.0001 \\
    \midrule
    
    \textbf{W2S guidance} & & & & & & & & \\
    \quad Guidance Scale \(w\) & / & 0.3 & 0.3 & 0.5 & 0.6 / 0.5 (REPA) & / & 0.3 & 0.3 \\
    \quad Inferior model & / & SiT-S/2 (T/4) & / & / & / & / & SiT-S/2 (T/4) & / \\
    \quad Output layers & / & / & 4 (/12) & / & 4 (/12) & / & / & 4 (/12) \\
    \quad Total params & 137.8M & 137.8 (+39.5)M & 139.0M & 137.8M & 139.0M & 137.8M  & 137.8 (+39.5)M & 139.0M \\
    \quad time/it & 1.00 & 1.27 & 1.02 & 1.23 & 1.22 & 1.00 & 1.26 & 1.02 \\
    \quad Segmented stamp \(\tau\) & / & / & / & / & 0.2 & / & / & / \\

    \bottomrule
    \end{tabular}
    }
    
    \caption{Training information for W2S guidance experiments. All models are trained on SiT-B/2. For AG, we train a separate weak model with T/4 iterations, where T is the strong model's iteration.}
    
    \label{tab:w2s_training_config}
\end{table*}

\noindent Besides ablations on the segmentation timestamp $\tau$ provided in the main paper, here we provide an additional ablation on the inference-time guidance scale $w$. We fixed the segmented timestep \(\tau=0.5\) and ablate the guidance scale of CFG and SLG. As illustrated in~\Cref{fig:supp_ablation_grid_laion,fig:supp_ablation_grid_mscoco}, this analysis highlights a notable trade-off: CFG excels at semantic adherence (measured by HPSv2.1), but its aesthetic scores are comparatively low.  Conversely, SLG produces high aesthetic quality but remains less competitive on HPSv2.1. Our method, SGG, successfully synergizes these two, achieving strong, comparable results across both metrics.

\subsection{More metrics on condition-adherence}
\label{app:cond_adherence}

\noindent\textbf{CLIPScore and GenEval}. Besides HPSv2.1~\cite{wu2023human}, we additionally report CLIPScore~\cite{hessel2021clipscore} and GenEval~\cite{ghosh2023geneval} on MS-COCO-1K and LAION-5B-1K for SD3.5-medium. As shown in \Cref{tab:sd35_guidance_subset}, SGG improves Aesthetic while remaining competitive on condition-based metrics.

\begin{table*}[ht]
    \centering
    \setlength{\tabcolsep}{5pt}
    \renewcommand{\arraystretch}{1.08}
    \small
    \resizebox{0.98\textwidth}{!}{%
    \begin{tabular}{l|cccc|cccc}
        \toprule
        \textbf{Method} & \multicolumn{4}{c|}{\textbf{MS-COCO-1K}} & \multicolumn{4}{c}{\textbf{LAION-5B-1K}} \\
        \cmidrule(lr){2-5} \cmidrule(lr){6-9}
        & \textbf{HPSv2.1} & \textbf{Aesthetic} & \textbf{CLIP} & \textbf{GenEval}
        & \textbf{HPSv2.1} & \textbf{Aesthetic} & \textbf{CLIP} & \textbf{GenEval} \\
        \midrule
        CFG & 29.219 & 5.279 & 26.316 & 0.628 & 28.234 & 5.203 & 27.521 & \textbf{0.674} \\
        SLG & 27.295 & 5.714 & 25.145 & 0.507 & 26.050 & 5.512 & 25.440 & 0.523 \\
        CFG+SLG & 28.931 & 5.678 & 26.485 & 0.598 & 27.246 & 5.421 & 27.154 & 0.638 \\
        CFG (linear schedule) & 28.833 & 5.261 & 26.351 & \underline{0.633} & 27.954 & 5.220 & \underline{27.777} & 0.665 \\
        SGG (swap orders) & 27.393 & 5.366 & 25.523 & 0.534 & 26.270 & 5.311 & 26.283 & 0.579 \\
        \rowcolor{blue!10}
        SGG (w/o CDG (CFG)) & 26.050 & 5.685 & 24.630 & 0.497 & 25.195 & 5.516 & 24.800 & 0.515 \\
        \rowcolor{blue!10}
        SGG (w/o CAG (SLG)) & 27.832 & 5.207 & 26.112 & 0.602 & 27.173 & 5.178 & 27.422 & 0.647 \\
        \rowcolor{blue!10}
        SGG & \textbf{29.736} & \underline{5.717} & \textbf{26.713} & 0.632 & \textbf{28.687} & \underline{5.518} & 27.649 & 0.668 \\
        \rowcolor{blue!10}
        SGG (linear schedule) & \underline{29.712} & \textbf{5.752} & \underline{26.595} & \textbf{0.637} & \underline{28.564} & \textbf{5.525} & \textbf{27.783} & \underline{0.672} \\
        \bottomrule
    \end{tabular}%
    }
    \vspace{-0.2cm}
    \caption{SD3.5-medium on MS-COCO-1K and LAION-5B-1K}
    \vspace{-0.2cm}
    \label{tab:sd35_guidance_subset}
\end{table*}

\subsection{Ablations and guidance schedules}
\label{app:abl_schedule}

\noindent\textbf{Guidance schedules.} SGG is orthogonal to scalar guidance scheduling $w(t)$~\cite{malarz2025classifier}: it changes the guidance \emph{family} across noise regimes and can be combined with standard schedules. We include linear $w(t)$ variants for CFG and SGG in \Cref{tab:sd35_guidance_subset}, with mild early-time clamping (e.g., starting from a non-trivial scale), increasing schedules do not weaken conditioning.

\noindent\textbf{More ablations.} We test (i) swapping CDG/CAG orders, (ii) removing CDG (CFG) from SGG, and (iii) removing CAG (SLG) from SGG. All variants are inferior to standard SGG in \Cref{tab:sd35_guidance_subset}, supporting the intended regime assignment.

\subsection{Extension to video generation}

To further evaluate the applicability of Segmented Guidance (SGG) principle across modalities, we extend our experiments to video generation using the \texttt{Wan2.1-1.3B}~\cite{wan2025wan} model on the subset of VBench~\cite{huang2024vbench} prompts corresponding to the metrics. For inference configuration, we adhere to the default setting with 50 sampling steps and 5.0 CFG scale. For SGG, we set the segmented timestamp \(\tau=25\), and SLG scale 3.0. We selected 6 metrics and calculate the average score.
The quantitative results in Table~\ref{tab:supp_wan} demonstrate that SGG manages to generate videos with better Aesthetic and imaging quality, while also remain competitive on physical plausibility.

\section{Training implementation and analysis}

\subsection{Training time settings}

\noindent\textbf{Models and metrics selection}. We conduct training evaluation mainly on SiT-B/2 model~\cite{ma2024sit} due to computational constraints. We use lognormal-timestep sampling throughout all experiments to boost convergence, following~\cite{shin2025deeply}. We perform experiments in both unconditional and conditional settings. CAG methods are applied in both settings, whereas CDG method is naturally applied only in conditional training. For the conditional setting. All models are trained for 400k iterations. The sampling configuration is SDE Euler-Maruyama
 sampler with steps=250. 
We report the FID, sFID and Inception Score for all methods.

\noindent\textbf{Implementation details of training W2S variants}.
Here we summarize the training-time implementation of AG, BR, CFG (our reimplementation of MG), and SGG.
The objective of the weak model in all variants shares the same base regression target \(\mathbf{u}=\epsilon - \mathbf{x}_0\)
and differs only in how the weak prediction and the W2S-modified strong target are constructed.
Further hyperparameter choices and information are listed in~\Cref{tab:w2s_training_config}.

\noindent\textbf{AG}.
For autoguidance~\cite{karras2024guiding}, we use a separate weak model $\mathbf{v}_{\theta_{\mathrm{w}}}$ with a smaller backbone (SiT-S/2) and a strong model $\mathbf{v}_\theta$ (SiT-B/2).
The weak model is updated once every 4 updates of the strong model.

\noindent\textbf{BR}.
In BR, the weak prediction is implemented as a shallow branch head
$\mathbf{v}^{\mathrm{br}}_\theta(\mathbf{x}_t, t, \mathbf{c})$ (same architecture of \texttt{FinalLayer()}) that taps into intermediate features of the same transformer, while the final head
$\mathbf{v}^{\mathrm{full}}_\theta(\mathbf{x}_t, t, \mathbf{c})$ serves as the strong output. The extra computational overhead of this model is negligible ( \(2\%\), \ie, time/it = 1.02)

\noindent\textbf{MG (CFG reimplementation).}
For MG/CFG, the weak prediction is provided by the unconditional branch
$\mathbf{v}_\theta(\mathbf{x}_t, t, \emptyset)$ of the same model, while
$\mathbf{v}_\theta(\mathbf{x}_t, t, c)$ is the strong (conditional) output. We keep the default \texttt{DROPOUT} rate 0.1 to train the unconditional model, as CFG~\cite{ho2021classifier}. This is equivalent to Model Guidance~\cite{tang2025diffusion} at the parameterization level.

\noindent\textbf{SGG}.
SGG combines a condition-agnostic weak signal (BR) and a condition-dependent weak signal (CFG) through a time-dependent switch. With segmented timestamp set to \(\tau=0.2\). (Inspired by the inference time setting of~\cite{yurepresentation} on ImageNet, we also apply guidance interval~\cite{Kynkaanniemi2024} from \([0.8,0.2]\) to avoid extreme high noise level  experiments for SGG) The overall architecture is identical to BR while keeping the conditional/unconditional training style to create CFG signal. The extra parameter overhead is \(0.8\%\) compared to vanilla SiT model. Further details can be referred in~\Cref{tab:w2s_training_config}.


\subsection{Training instability of SLG}

\begin{figure}[t]
    \centering
    \includegraphics[width=0.95\linewidth]{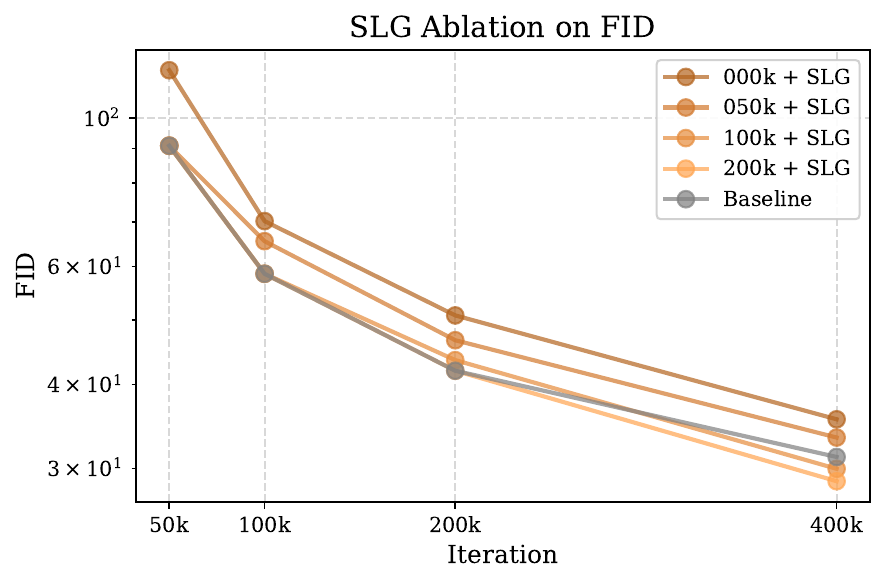}
    \caption{SLG in training, applied in different iterations.}
    \vspace{-0.5em}
    \label{fig:supp_slg_ablation}
\end{figure}

Compared to AG, BR, and MG, we observe that applying Skip Layer Guidance (SLG) during training from scratch exhibits degradation. We attribute this to the high variance of the synthetic weak signal generated by layer perturbation when applied to an \textit{unconverged} model. To address this, we use a warm-up phase utilizing pure regression loss. As illustrated in~\Cref{fig:supp_slg_ablation}, extending this warm-up period improves performance, eventually surpassing the baseline at 100k and 200k iterations, likely by ensuring the model is robust enough to provide a stable weak signal.
\begin{table*}[b]
    \centering
    \resizebox{0.98\textwidth}{!}{%
    \setlength{\tabcolsep}{12pt} 
    \renewcommand{\arraystretch}{1.1}
    
    \begin{tabular}{llcccccc}
    \toprule
    & & \multicolumn{3}{c}{Conditional} & \multicolumn{3}{c}{Unconditional} \\
    \cmidrule(lr){3-5} \cmidrule(lr){6-8}
    Method & Guidance & FID $\downarrow$ & sFID $\downarrow$ & Inception Score $\uparrow$ & FID $\downarrow$ & sFID $\downarrow$ & Inception Score $\uparrow$ \\
    \midrule
    Baseline & \(w=0.0\) & 31.22 & 6.41 & 49.59 & 61.27 & 7.00 & 17.33 \\
    \midrule
    \multirow{3}{*}{BR} & \(w=0.2\) & 18.33 & 4.71 & 70.45 & 46.37 & 4.93 & 20.23 \\
    & \(w=0.3\) & 16.02 & 5.13 & 76.21 & 43.25 & 5.11 & 20.66 \\
    & \(w=0.4\) & 15.09 & 7.64 & 80.29 & 42.13 & 7.59 & 20.43 \\
    \midrule
    \multirow{3}{*}{AG} & \(w=0.2\) & 18.71 & 4.97 & 73.30 & 50.97 & 5.52 & 19.40 \\
    & \(w=0.3\) & 13.96 & 4.68 & 88.35 & 45.97 & 4.94 & 20.32 \\
    & \(w=0.4\) & 10.91 & 5.40 & 102.89 & 42.44 & 5.31 & 21.01 \\
    \bottomrule
    \end{tabular}
    }
    \caption{Ablation study on guidance scale ($w$) for W2S training methods (BR and AG) in both conditional and unconditional settings. All models are SiT-B/2 trained on ImageNet 256x256.}
    \label{tab:w2s_training_ablation}
\end{table*}
Despite the marginal performance gains, the limitations of this approach are more obvious. Pure SLG~\cite{hyung2025spatiotemporal} necessitates an additional forward pass similar to CFG~\cite{ho2021classifier}, yet the resulting weak signal is often inferior to the unconditional output. Furthermore, tuning the warm-up hyperparameter becomes computationally prohibitive when scaling to larger tasks. Given these unfavorable trade-offs, we thus exclude SLG from our proposed training framework.

However, for well-trained models at scale~\cite{esser2024scaling}, the utility of SLG becomes apparent. Training a separate inferior model for AutoGuidance~\cite{karras2024guiding} is often impractical for large-scale architectures like Stable Diffusion~\cite{rombach2022high,esser2024scaling}. In these scenarios, where the primary model is sufficiently robust, self-degradation techniques like SLG provide an efficient mechanism for constructing the Condition-Agnostic Guidance (CAG) signal~\cite{hyung2025spatiotemporal,ahn2024self}.

\subsection{Ablations on training guidance scale}

Ablations on segmented timestamp \(\tau\) of SGG is provided in the main paper, here we provide more ablations on the selection of training time guidance scale \(w\) on AG and BR variants on both conditional and unconditional setting, as illustrated in~\Cref{tab:w2s_training_ablation}.

\section{Discussion}

\noindent\textbf{The generalization trade-off.} Theoretically, a velocity predictor \(\dot{\mathbf{v}}\) that perfectly minimizes the objective is capable of faithfully reconstructing training data points, a state characterized as memorization~\cite{gumemorization}. In practice, however, network inductive biases and inevitable approximation errors prevent this, instead enabling the model to generalize to unseen data~\cite{kadkhodaie2024generalizationdiffusionmodelsarises}. Yet, when scaled to complex text-to-image tasks~\cite{esser2024scaling,rombach2022high}, these accumulated errors often cause the unguided generation trajectory to diverge from the perceptually acceptable manifold—a deviation that often persists regardless of the number of sampling steps. Consequently, guidance techniques are required to steer the trajectory back toward perceptually acceptable regions, albeit at the cost of additional computation (\eg, computing an extra weak signal). Thus, the generalization capability of diffusion models presents trade-offs: it is simultaneously enabled by, yet suffers from, the approximation errors accumulated across sampling steps.

\noindent\textbf{More intuition on the proposed method}. 
CDG derives its guidance signal from an \emph{external} semantic discrepancy (i.e., \(\mathbf{c}\) vs.\ \(\emptyset\)), and thus primarily steers global content such as semantics, coarse structure, and layout. These attributes are largely determined at earlier denoising stages, where the model establishes low-frequency components of the sample. 
In contrast, CAG is driven by the model’s \emph{internal} prediction error under the condition, making its signal inherently condition-aligned and more effective for intra-class refinement, including local details and texture that emerge in later timesteps (high-frequency components)~\cite{karras2024guiding}. 
Consequently, SGG adopts a natural division of labor: it applies CDG in the high-noise regime to quickly locate the correct conditional manifold, and then switches to CAG in the low-noise regime to refine fine-grained details while maintaining prompt consistency.

\clearpage
\section{More qualitative results.}
\subsection{Qualitative results on text-to-video models}

\begin{figure*}[b]
    \centering
    \includegraphics[width=0.85\linewidth]{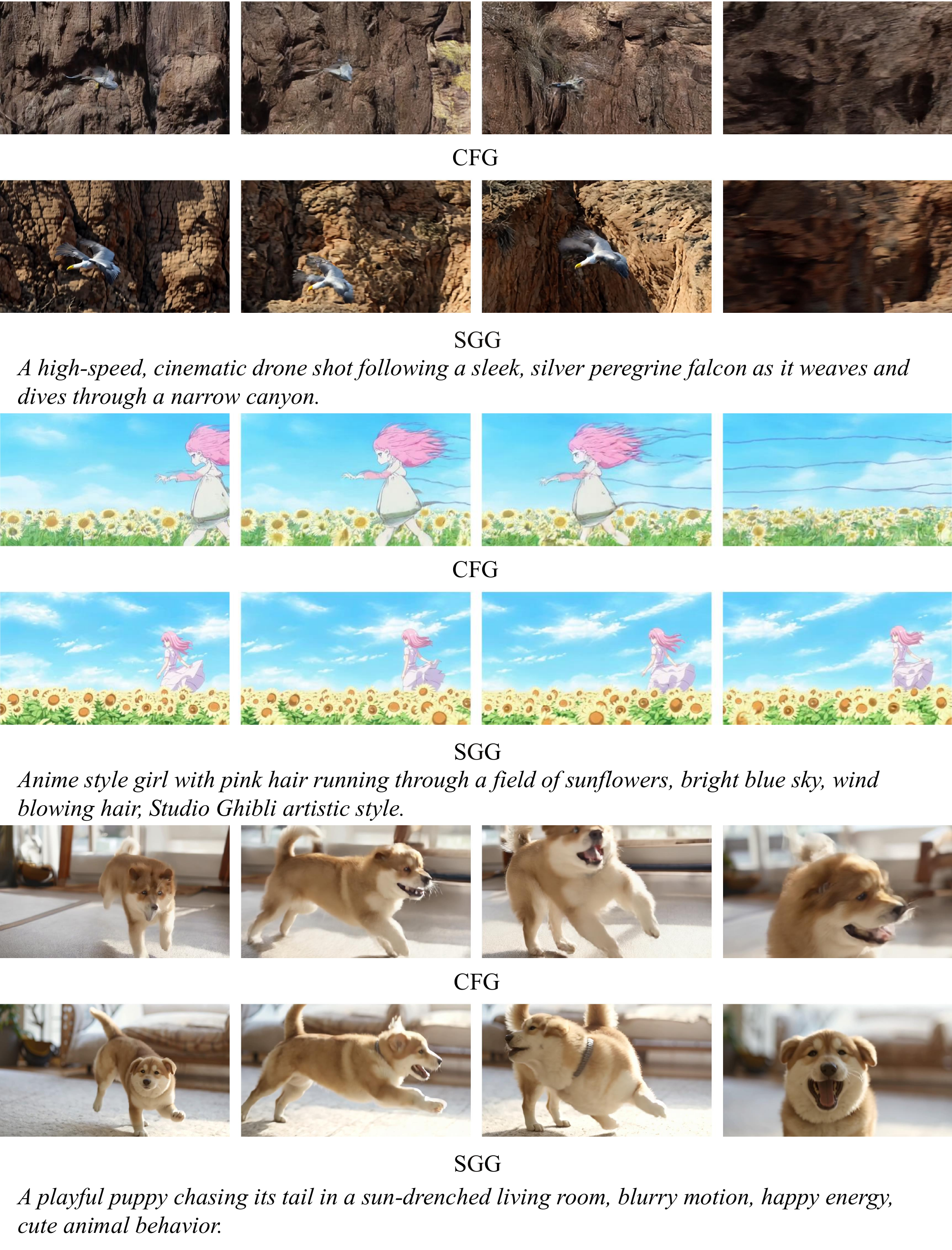}
    \caption{Qualitative comparison of CFG and SGG (Ours) on video generation}
    \label{fig:supp_ablation_grid_wan}
\end{figure*}

\clearpage
\subsection{Qualitative results on text-to-image models}
\begin{figure*}[b]
    \centering
    \includegraphics[width=\linewidth]{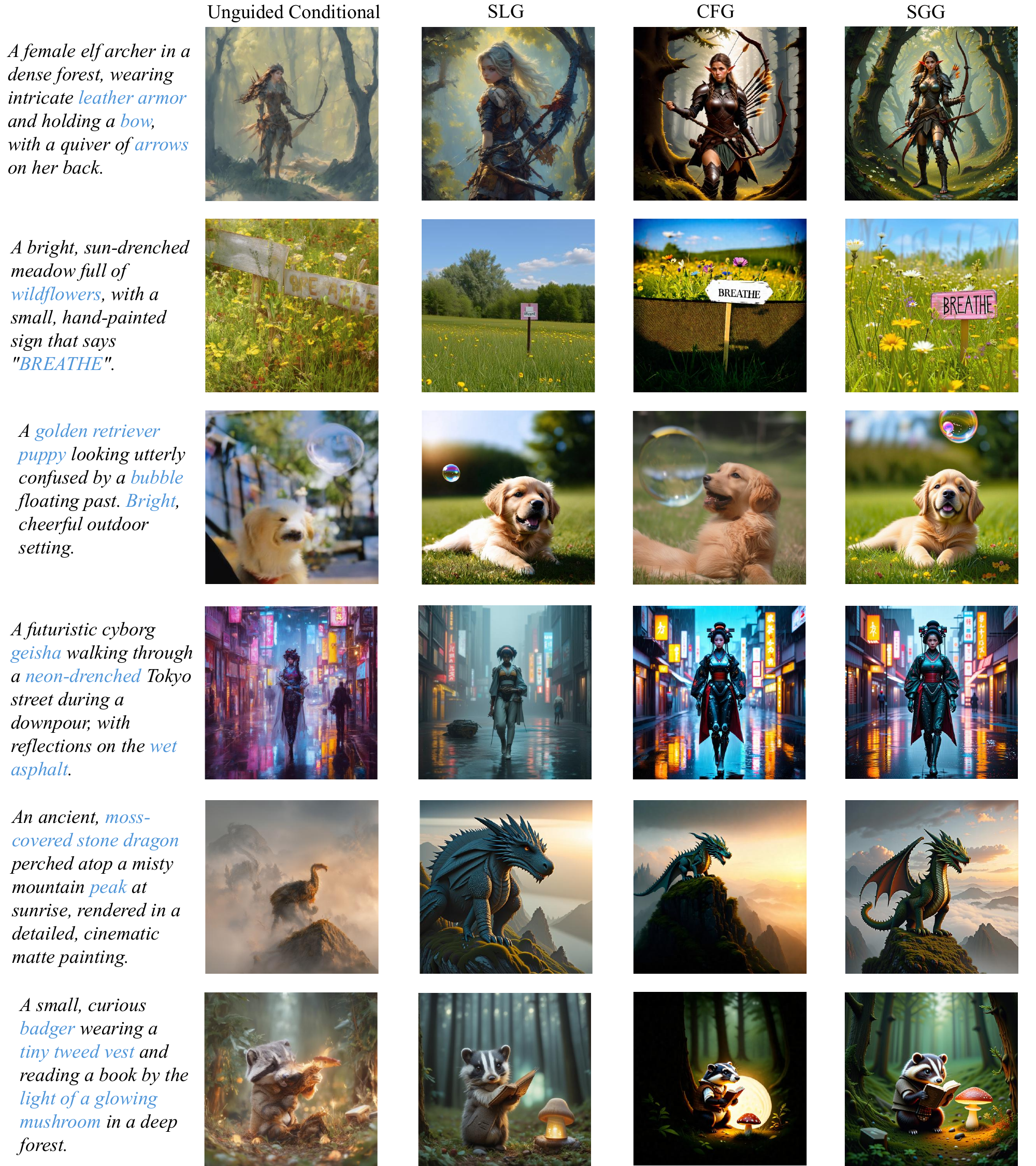}
    \caption{Qualitative Comparison between Unguided Conditional, CFG, SLG and SGG (Ours) (1/2)}
    \label{fig:supp_sd3_1}
\end{figure*}

\begin{figure*}[b]
    \centering
    \includegraphics[width=\linewidth]{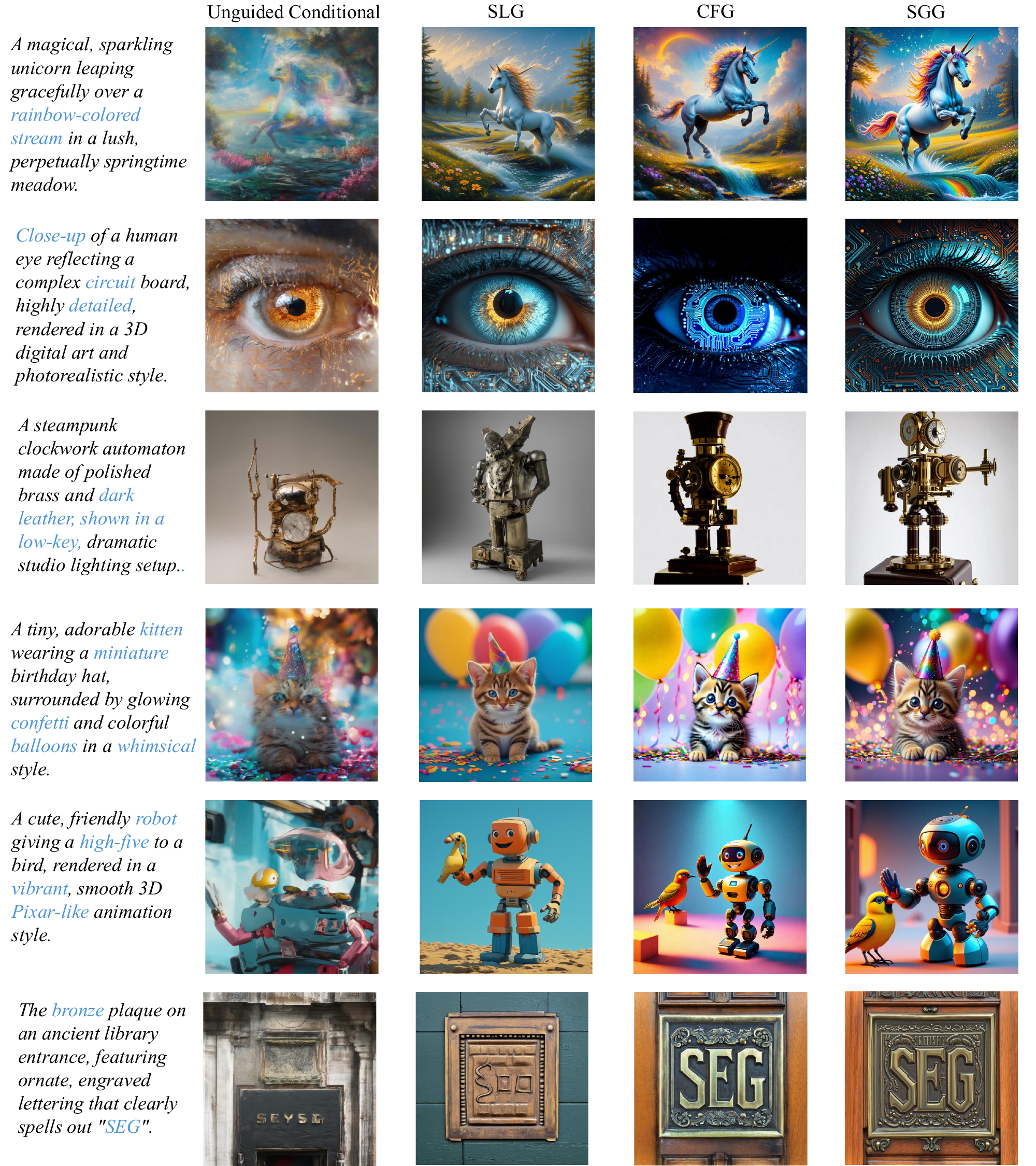}
    \caption{Qualitative Comparison between Unguided Conditional, CFG, SLG and SGG (Ours) (2/2)}
    \label{fig:supp_sd3_2}
\end{figure*}

\clearpage
\subsection{Qualitative results on ImageNet256}

\begin{figure*}[b]
    \centering
    \includegraphics[width=0.8\linewidth]{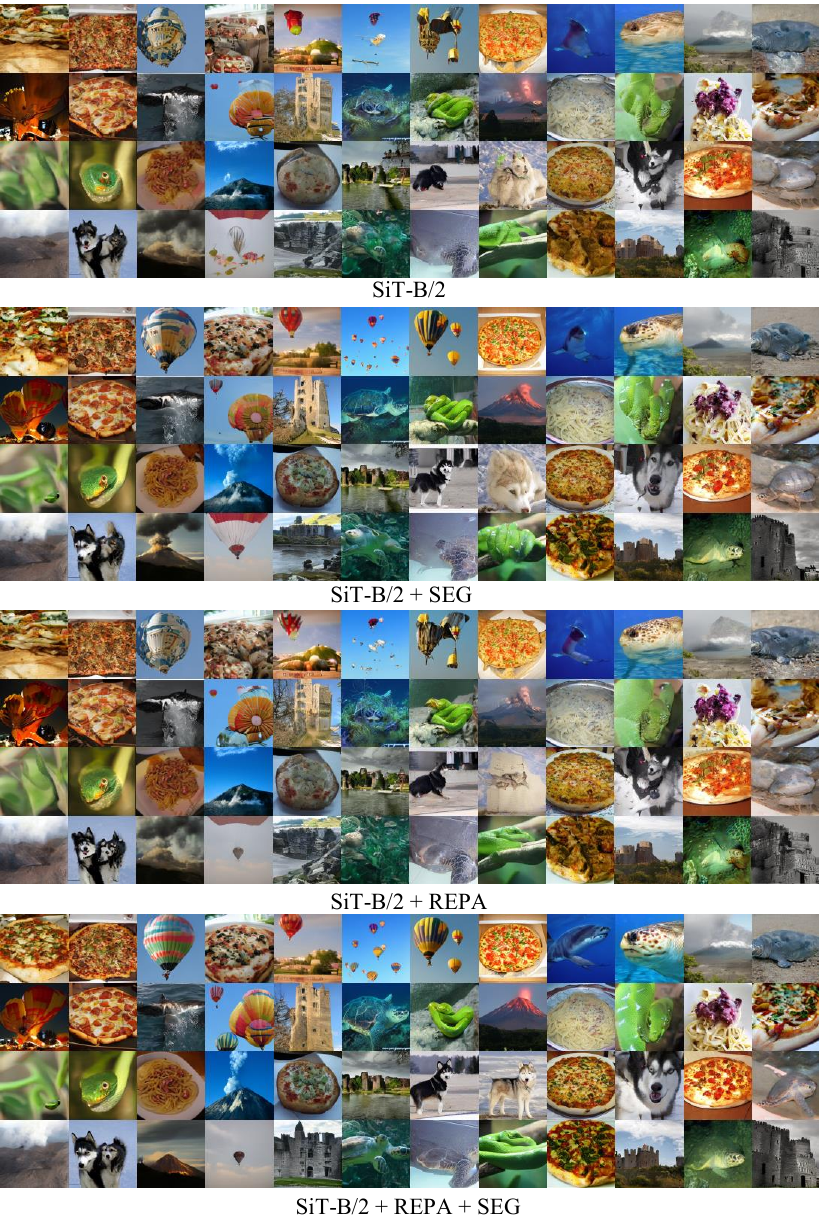}
    \caption{Qualitative Comparison between SiT-B/2 (Baseline), SGG (Ours), REPA, REPA+SGG}
    \label{fig:supp_ablation_grid_imagenet256}
\end{figure*}

\ifarxiv
\else
      {\small
    \bibliographystyle{ieeenat_fullname}
    \bibliography{sections/11_references}
    }
\fi

 \fi

\end{document}